\documentclass[10pt,twocolumn,letterpaper]{article}

\usepackage[pagenumbers]{cvpr} 

\usepackage{graphicx}
\usepackage{amsmath}
\usepackage{amssymb}
\usepackage{booktabs}

\usepackage{subfiles}
\usepackage{graphicx}
\usepackage{amsmath}
\usepackage{amssymb}
\usepackage{multirow}
\usepackage{makecell}
\usepackage{color}
\usepackage{bm}
\usepackage{marvosym}
\usepackage{xcolor}

\usepackage[accsupp]{axessibility} 

\newcommand\blfootnote[1]{%
    \begingroup
    \renewcommand\thefootnote{}\footnote{#1}%
    \addtocounter{footnote}{-1}%
    \endgroup
}

\newcommand{\FIG}[1]{\cref{#1}}
\newcommand{\TABLE}[1]{\cref{#1}}
\newcommand{\EQUA}[1]{\cref{#1}}

\newcommand{\MODEL}{Unnamed Model}

\newcommand{\MARK}[1]{\textcolor{black}{\textbf{#1}}}
\newcommand{\xmark}{$\times$}
\newcommand{\xm}{\xmark}
\newcommand{\cm}{\checkmark}

\newcommand{\HARDINDENT}{\hspace{1pc}}

\newcommand{\NA}{N/A}
\newcommand{\ETAL}{{\emph{et al.}}}

\newcommand{\IE}{{\emph{i.e.}}}

\newcommand{\SOCIALLSTM}{Social-LSTM}
\newcommand{\SOCIALLSTMCITE}{\cite{alahi2016social}}
\newcommand{\SOCIALGAN}{S-GAN}
\newcommand{\SOCIALGANCITE}{\cite{gupta2018social}}

\newcommand{\SRLSTMCITE}{\cite{zhang2019sr}}

\newcommand{\PEEKING}{Next}
\newcommand{\PEEKINGCITE}{\cite{liang2019peeking}}

\newcommand{\PECNET}{PECNet}
\newcommand{\PECNETCITE}{\cite{mangalam2020not}}
\newcommand{\SRLSTM}{SR-LSTM}

\newcommand{\STAR}{STAR}
\newcommand{\STARCITE}{\cite{yu2020spatio}}

\newcommand{\TRAJECTRONPP}{Trajectron++}
\newcommand{\TRAJECTRONPPCITE}{\cite{salzmann2020trajectron}}
\newcommand{\YNET}{$\textsf{Y}$-net}
\newcommand{\YNETCITE}{\cite{mangalam2020s}}

\newcommand{\STGCNN}{Social-STGCNN}
\newcommand{\STGCNNCITE}{\cite{mohamed2020social}}

\newcommand{\AGENTFORMER}{Agentformer}
\newcommand{\AGENTFORMERCITE}{\cite{yuan2021agentformer}}
\newcommand{\DAGNET}{DAG-Net}
\newcommand{\DAGNETCITE}{\cite{monti2021dag}}
\newcommand{\STCNET}{STC-Net}
\newcommand{\STCNETCITE}{\cite{li2021spatial}}
\newcommand{\MSN}{MSN}
\newcommand{\MSNCITE}{\cite{wong2021msn}}

\newcommand{\VMODEL}{V$^2$-Net}
\newcommand{\VCITE}{\cite{wong2022view}}
\newcommand{\MID}{MID}
\newcommand{\MIDCITE}{\cite{gu2022stochastic}}
\newcommand{\SHENET}{SHENet}
\newcommand{\SHENETCITE}{\cite{meng2022forecasting}}

\newcommand{\NSPSFM}{NSP-SFM}
\newcommand{\NSPSFMCITE}{\cite{yue2022human}}

\newcommand{\EVMODEL}{E-V$^2$-Net}
\newcommand{\EVCITE}{\cite{wong2023another}}
\newcommand{\EQMOTION}{EqMotion}
\newcommand{\EQMOTIONCITE}{\cite{xu2023eqmotion}}
\newcommand{\LED}{LED}
\newcommand{\LEDCITE}{\cite{mao2023leapfrog}}
\newcommand{\FLOWCHAIN}{FlowChain}
\newcommand{\FLOWCHAINCITE}{\cite{maeda2023fast}}
\newcommand{\IMP}{IMP}
\newcommand{\IMPCITE}{\cite{shi2023representing}}

\renewcommand{\MODEL}{SocialCircle}

\definecolor{cvprblue}{rgb}{0.21,0.49,0.74}
\usepackage[pagebackref,breaklinks,colorlinks,citecolor=cvprblue]{hyperref}

\usepackage[capitalize]{cleveref}
\crefname{section}{Sec.}{Secs.}
\Crefname{section}{Section}{Sections}
\Crefname{table}{Table}{Tables}
\crefname{table}{Tab.}{Tabs.}

\def\paperID{1945} 
\def\confName{CVPR}
\def\confYear{2024}

\title{
    SocialCircle: Learning the Angle-based Social Interaction Representation \\
    for Pedestrian Trajectory Prediction
}

\author{
    Conghao Wong$^{1*}$\quad
    Beihao Xia(\Letter)$^{1*}$\quad
    Ziqian Zou$^1$\quad
    Yulong Wang$^2$\quad
    Xinge You$^{1,3}$\\
    $^1$Huazhong University of Science and Technology\quad
    $^2$Huazhong Agricultural University \\
    $^3$Research Institute of Huazhong University of Science and Technology in Shenzhen \\
    {\tt\footnotesize \{conghaowong, xbh\_hust, ziqianzoulive\}@icloud.com,
    ylwang@mail.hzau.edu.cn,
    youxg@mail.hust.edu.cn}
}

\begin{document}
\maketitle


\begin{abstract}

Analyzing and forecasting trajectories of agents like pedestrians and cars in complex scenes has become more and more significant in many intelligent systems and applications.
The diversity and uncertainty in socially interactive behaviors among a rich variety of agents make this task more challenging than other deterministic computer vision tasks.
Researchers have made a lot of efforts to quantify the effects of these interactions on future trajectories through different mathematical models and network structures, but this problem has not been well solved.
Inspired by marine animals that localize the positions of their companions underwater through echoes, we build \textbf{a new angle-based trainable social interaction representation}, named \textbf{\MODEL}, for continuously reflecting the context of social interactions at different angular orientations relative to the target agent.
We validate the effect of the proposed \MODEL~by training it along with several newly released trajectory prediction models, and experiments show that the \MODEL~not only quantitatively improves the prediction performance, but also qualitatively helps better simulate social interactions when forecasting pedestrian trajectories in a way that is consistent with human intuitions.

\end{abstract}


\section{Introduction}

\blfootnote{
    * equal contribution.
    Codes are available at \url{https://github.com/cocoon2wong/SocialCircle}.
}

\begin{figure}[t]
    \centering
    \includegraphics[width=1.0\linewidth]{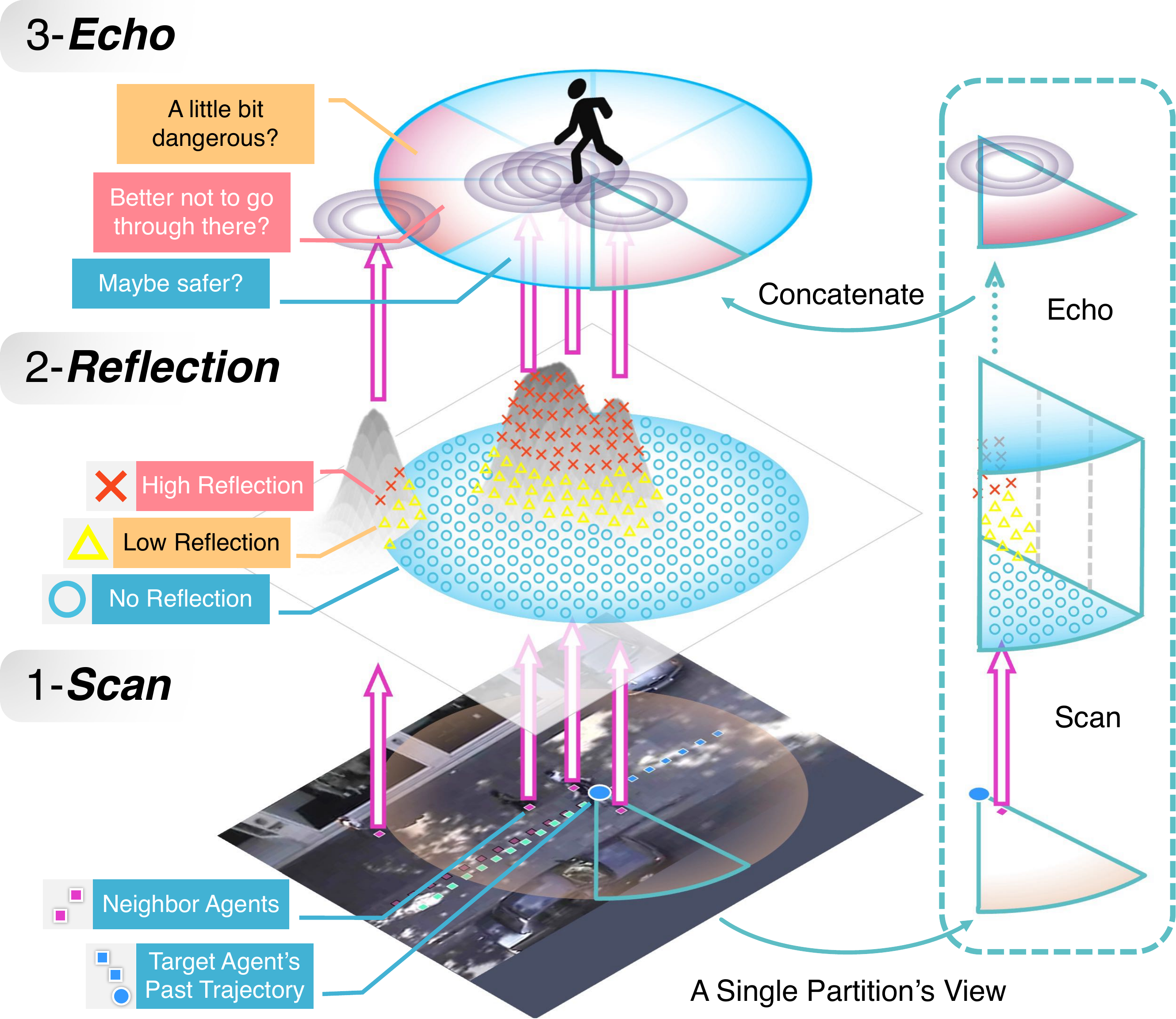}
    \caption{
        Motivation Illustration.
        Analogous to marine animals like dolphins and whales localizing other companions underwater through echolocation, we analyze agents' reactions to the potential socially interactive behaviors by assuming (1) they first \textbf{\emph{Scan}} their interaction environment by sending signals from all angles, (2) then all neighbors feedback their \textbf{\emph{Reflection}} signals to tell their directions, and (3) finally the target agent could make interactive decisions by the received \textbf{\emph{echoes}} at different \textbf{\emph{angular}} orientations.
    }
    \label{fig_intro}
\end{figure}

Analyzing, understanding, and forecasting behaviors of intelligent agents have been significantly required by more and more intelligent systems and applications.
Due to the ease of access and analysis of trajectories, analyzing agents' behaviors through trajectories has also become a common approach.
Trajectory prediction aims at forecasting agents' all possible future trajectories during a specific period by taking into account the positions of all agents that appeared in the scene \cite{alahi2016social}.
It also considers the potential interactive behaviors \cite{xu2022socialvae,shi2022social,kothari2021interpretable,amirian2019social,gupta2018social} as well as the scene constraints \cite{xue2020scene,sadeghian2019sophie,lisotto2019social,chen2022scept,meng2022forecasting,barata2021sparse,xia2022cscnet,ngiam2021scene} when making predictions.

The \textbf{social interaction} \cite{alahi2016social,pellegrini2009youll} (also known as the \textbf{agent-to-agent interaction}) considered in trajectory prediction takes into account not only all kinds of interactive behaviors among different agents but how they affect their trajectories.
Current social-interaction-modeling methods in the trajectory prediction task can be classified roughly into \emph{Model-based} and \emph{Model-free} two classes\cite{yue2022human}.
\emph{Model-based} methods may take some particular ``rules'' \cite{yue2022human} as the primary foundation for the prediction.
For example, the Social-Force-based methods \cite{helbing1995social,pellegrini2009youll} model and simulate agents' behaviors mainly according to the rules in Newtonian mechanics.
Some other methods like \cite{xie2017learning,yue2022human,barata2021sparse} also turn trajectory prediction into an optimization problem by introducing their different mathematical models.
However, designing a generalized ``rule'' that fits most socially interactive cases is often difficult, making them challenging to apply to complex scenes.
On the contrary, \emph{model-free} methods are mostly driven by data, and few manual interventions are considered.
For example, graph-based methods \cite{su2022trajectory,li2022intention,cao2021spectral} may build a series of spatial or temporal graph structures, thus learning to simulate agents' social interactions.
Most model-free methods could fully utilize the ability of neural networks to fit data, but they may heavily rely on different network structures and pose limited explainability.

``Rules'' and ``data'' play essential roles but put different limitations on these methods accordingly.
A natural thought is to add several ``lite-rules'' to data-driven backbones to provide limited constraints as guidance to improve either the data-fit process or the explainability.
In short, we want to constrain the learning process with relatively weak rules for social interactions rather than solid mathematical rules, thus benefiting from both rules and data-fit capabilities.

Analyzing agents' interactive behaviors through bionics and psychology is a natural choice.
Animals would not analyze others' behaviors by solving complex equations but with relatively simple judgment rules when planning trajectories.
Some researchers in the social psychology area also point out that each agent in a complex multiagent system tends to behave and interact with each other according to simple rules rather than extensive computations, which inspired a series of agent-based simulation models that have been widely applied in economics and political science \cite{smith2007agent}.
It is fascinating that some marine animals can locate others in the deep sea through \emph{echolocation} rather than visual factors due to the weak light.
They may firstly \textbf{scan} the environment by sending some unique signals (like ultrasounds) at different angles, which could be \textbf{reflected} in contact with others and produce \textbf{echoes}.
Then, they gather echoes from different directions, thus locating, interacting, or communicating with others, and finally modifying their behaviors.

As shown in \FIG{fig_intro}, the echolocation is similar to how agents interact with others.
Only a few ``rules'' are established, like the time from they send to receive the echo and the direction where it comes.
This way, we bring a simple priori to model social behaviors that interactions are considered to be \textbf{angle-based}.
In detail, all interactive behaviors are considered in a special angle space where the angle $\theta$ (\emph{which direction the echo comes from}) plays as the independent variable.
We assume that most social interactions can be ``inferred'' by several simple components corresponding to each angle $\theta$, like the velocity of each participant (\emph{in which way the participant's position changes during two echolocations}) and the distance between each participant and the target agent (\emph{how long the echo arrives since scanning}).
Thus, we can obtain an angle-based vector function $\mathbf{f}(\theta)~\left(0 \leq \theta < 2\pi \right)$ to represent the current socially interactive context when forecasting trajectories.
We call that angle-based social interaction representation \textbf{\MODEL}.

\MODEL~can be classified as \emph{Model-based}, but it is also inspired by model-free methods to fit data with relatively weak rules, \IE, simple components at different angles.
Then, the observed trajectory and the angle-based \MODEL~will be analyzed together in a data-driven way, as they could be both \emph{treated as} sequences, to catch the temporal-attentive portions in trajectories and the angle-attentive portions in the current interactive context simultaneously, thus establishing connections among these weak rules and agents' real-world socially interactive behaviors as well as the forecasted trajectories.

In summary, we contribute
(1) The angle-based \MODEL~representation for pedestrian trajectory prediction to model social interactive behaviors;
(2) The serialized modeling strategy that treats and encodes the spatial social interactions in the temporal sequences' way along with trajectories;
(3) Experiments on multiple backbone prediction models show quantitative and qualitative superiority.

\section{Related Work}


\HARDINDENT\textbf{Model-based Social Interaction Methods.}
Model-based methods aim to use mathematical rules as the foundation to forecast trajectories.
The classic Social Force Model \cite{helbing1995social} is proposed to model human dynamics with Newtonian mechanics.
Pellegrini \ETAL~\cite{pellegrini2009youll} introduce Social Force factors to model social behaviors in the multi-agent tracking task.
More Social-Force-based methods like \cite{zanlungo2011social,luber2010people,mehran2009abnormal} are also proposed to model crowds' interactions.

Other mathematical tools and models are also used to simulate socially interactive behaviors when forecasting trajectories.
Xie \ETAL~\cite{xie2017learning} propose the ``Dark Matter'' model to simulate and forecast social behaviors with fields and agent-based Lagrangian Mechanics.
Xia \ETAL~\cite{xia2022cscnet} propose a social transfer function to model human socially interactive behaviors via a uniform way across multiple prediction scenes.
Yue \ETAL~\cite{yue2022human} propose a neural differential equation model, in which the explicit physics model serves as a strong inductive bias in modeling pedestrian behaviors.

However, these methods are often difficult to cover all possible socially interactive cases.
Even though some methods like \cite{yue2022human,xia2022cscnet} draw on the advantages of data-driven approaches to make some key parameters trainable, they may still be limited by the complex mathematical rules and equations in complex prediction scenes.

\textbf{Model-free Social Interaction Methods.}
Model-free methods simulate interactive behaviors mostly through a data-driven form.
Alahi \ETAL~\cite{alahi2016social} propose the Social-Pooling method to connect nearby sequences to share hidden states with each other, thus simulating the information-sharing process.
Variations of social pooling methods like \cite{gupta2018social,sadeghian2019sophie} are proposed to pool features by considering different scales or locations simultaneously.
Grid-based methods like \cite{kothari2020human} have been proposed to explore additional simple rules to enhance the capacity of pooling methods.
With the quick development of graph neural networks, graph structures have been widely used to model social interactions.
Graph Attention Networks (GATs) \cite{monti2021dag,liu2021avgcn}, Graph Convolutional Networks \cite{su2022trajectory,shi2021sgcn,dang2021msr} are employed to simulate interactions as the edges between different nodes.

Most model-free methods prefer to focus more on the structure through which to fit the data so that the predicted trajectory could reflect the effects of social interactive cues.
In this process, few direct mathematical rules are constraints, making them more dependent on different network structures and high-quality data.

The proposed \MODEL~tries to address these problems by introducing ``lite-rules'' to these trainable backbones, thus taking advantage of the data-driven approach combined with the explainability of the model-based approach to model interactive behaviors.
It also avoids designing complex mathematical models or solving complex equations during the interaction modeling.

\section{Method}

\begin{figure*}[t]
    \centering
    \includegraphics[width=1.0\linewidth]{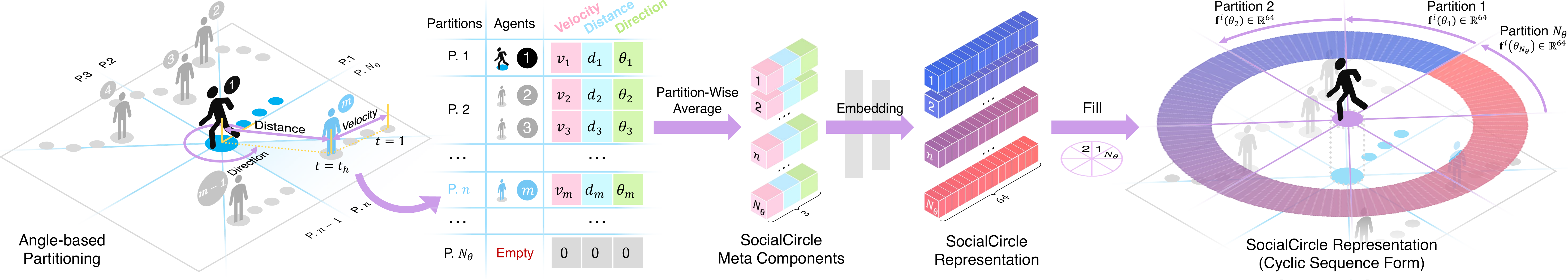}
    \caption{
        Computation pipeline of the proposed \MODEL.
        Each agent's \MODEL~is different to others'.
        For a target agent, it first computes three meta components: velocity, distance, and direction.
        Then, these meta components will be averaged within each angle-based \MODEL~partition, and finally embedded into the set of high-dimensional head-to-tail cyclic representation $\mathbf{f}^i\left(\theta_n\right)~\left(1 \leq n \leq N_\theta\right)$.
    }
    \label{fig_method}
\end{figure*}

\begin{figure}[t]
    \centering
    \includegraphics[width=1.0\linewidth]{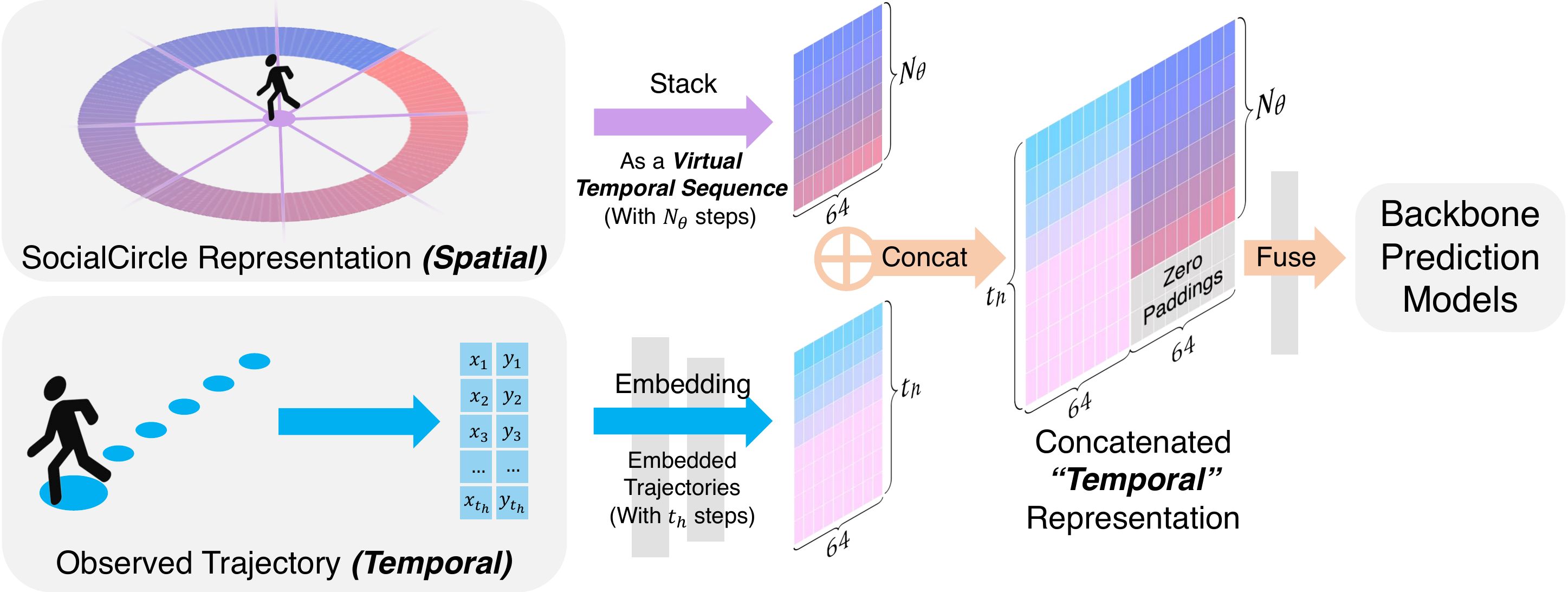}
    \caption{
        The serialized modeling of social interactions.
        \MODEL~is treated as a ``virtual'' temporal sequence that shares the same sequence shape as the embedded trajectory by adding zero paddings, so they could be fused to forecast trajectories together.
    }
    \label{fig_method_backbone}
\end{figure}

\HARDINDENT\textbf{Formulations.}
This work only concerns trajectories of 2D coordinates $\mathbf{p} = (x, y)^\top$.
Denote the historical trajectory of agent (pedestrian) $i$ during $t_h$ observation steps as $\mathbf{X}^i = \left(\mathbf{p}_1^i, ..., \mathbf{p}_{t_h}^i\right)^{\top}$, trajectory prediction focused in this paper aims at forecasting one or more possible future trajectories $\hat{\mathbf{Y}}^i = \left(\hat{\mathbf{p}}_{t_h + 1}^i, ..., \hat{\mathbf{p}}_{t_h + t_f}^i\right)^{\top}$ through its observed $\mathbf{X}^i$ and all its' neighbors' trajectories $\mathbf{\mathcal{X}}^{/i} = \left\{\mathbf{X}^j | 1 \leq j \leq N_a, j \neq i\right\}$ along with the scene image $\mathbf{I}_{t_h}$.

\textbf{Angle-based \MODEL~Representation.}
In this paper, all social-interaction-related operations are described and implemented in an ``angle'' space, where the angle $\theta$ is the independent variable that describes the location of interactive behaviors.
We first define the angle $\theta^i (j) \in \left[0, 2\pi\right)$ to represent the relative position of a neighbor agent $j$ to the target agent $i$.
It is computed as the ``direction'' of the vector that begins from agent $i$ and ends at agent $j$ at the current observation step ($t = t_h$).
Formally,
\begin{equation}
    \theta^i (j) = {\rm{atan2}}\left(\mathbf{p}^j_{t_h} - \mathbf{p}^i_{t_h}\right).
\end{equation}
Here, ${\rm{atan2}}$ is the ``quadrant-sensitive'' $\arctan$ function that computes angle of the input $\mathbf{p} = (x, y)^\top$ from $0$ to $2\pi$.

Agent $i$'s \textbf{\MODEL~representation} (short for \textbf{\MODEL}) can be treated as a head-to-tail cyclic vector function $\mathbf{f}^i(\theta)$ for all $\theta \in \left[0, 2\pi\right)$.
To make the computation easier, the angle variables will be discretized into $N_\theta$ ``partitions''.
This way, agent $i$'s \MODEL~can be denoted as
\begin{equation}
    \mathbf{f}^i = \left(
        \mathbf{f}^i \left(\theta_1\right),
        \mathbf{f}^i \left(\theta_2\right),
        ...,
        \mathbf{f}^i \left(\theta_{N_\theta}\right)
    \right)^{\top}.
\end{equation}
Here, $0 = \theta_0 < \theta_1 < ... < \theta_{N_\theta} = 2\pi$.
As shown in \FIG{fig_method}, each $\mathbf{f}^i \left(\theta_n\right) \in \mathbb{R}^{d_{\rm{sc}}}~\left(n = 1, 2, ..., N_\theta\right)$ is used to represent the overall interactive effort in the $n$th partition caused by all participants from the set $\mathbf{N}^i(\theta_n)$, which satisfies
\begin{equation}
    \theta_{n-1} \leq \theta^i (j) < \theta_n,~~\forall j \in \mathbf{N}^i(\theta_n).
\end{equation}
We treat agent $i$ as its self-neighbor in the 1st partition.
Denote the number of agents in $\mathbf{N}^i(\theta_n)$ as $\left| \mathbf{N}^i(\theta_n) \right|$, we have
\begin{align}
    \label{eq_selfneighbor}
i \in \mathbf{N}^i(\theta_1),~~
    \sum_n \left| \mathbf{N}^i(\theta_n) \right| = N_a.
\end{align}

\textbf{\MODEL~Meta Components.}
Each \MODEL~partition is computed via three meta components:
\begin{equation}
    \mathbf{f}^i_{\rm{meta}}\left(\theta_n\right) = 
        \left(
            \mathbf{f}^i_{\rm{vel}}(\theta_n), 
            \mathbf{f}^i_{\rm{dis}}(\theta_n), 
            \mathbf{f}^i_{\rm{dir}}(\theta_n)
        \right)^\top.
\end{equation}

\noindent (a) \textbf{Velocity $\mathbf{f}^i_{\rm{vel}}$.}
Agents with higher velocities may pose potentially more significant dangers to others around them.
\MODEL~takes the ``average velocity'' (the movement length during the observation period) of all neighbors in one partition to simulate this interactive factor.
Formally,
\begin{equation}
    \label{eq_vel}
    \mathbf{f}^i_{\rm{vel}}\left(\theta_n\right) =
        \frac{1}{\left| \mathbf{N}^i(\theta_n) \right|}
        \sum_{j \in \mathbf{N}^i(\theta_n)}
        \left\Vert \mathbf{p}^j_{t_h} - \mathbf{p}^j_1 \right\Vert _2,
\end{equation}

\noindent (b) \textbf{Distance $\mathbf{f}^i_{\rm{dis}}$.}
Agents present different interaction preferences as the distance to the interaction participant changes.
\MODEL~takes the average Euclidean distance (at $t = t_h$ moment) between the target agent and all its neighbors in one partition to model this factor.
Formally,
\begin{equation}
    \mathbf{f}^i_{\rm{dis}}\left(\theta_n\right) =
        \frac{1}{\left| \mathbf{N}^i(\theta_n) \right|}
        \sum_{j \in \mathbf{N}^i(\theta_n)}
        \left\Vert \mathbf{p}^i_{t_h} - \mathbf{p}^j_{t_h} \right\Vert _2.
\end{equation}

\noindent (c) \textbf{Direction $\mathbf{f}^i_{\rm{dir}}$.}
Partitioning the continuous $\theta \in \left[0, 2\pi\right)$ may cause the loss of angle details.
We use the average angles of all neighbors in one partition relative to the target agent as a compensation factor.
It also acts as a positional coding term to distinguish different partitions.
Formally,
\begin{equation}
    \mathbf{f}^i_{\rm{dir}}\left(\theta_n\right) =
        \frac{1}{\left| \mathbf{N}^i(\theta_n) \right|}
        \sum_{j \in \mathbf{N}^i(\theta_n)}
        \theta^i (j).
\end{equation}

\textbf{Serialized Modeling of Social Interaction.}
\MODEL~partitions $\left\{\mathbf{f}^i \left(\theta_n\right)\right\}_n$ are obtained by concatenating and embedding all meta components.
Formally,
\begin{equation}
    \mathbf{f}^i \left(\theta_n\right) = \left\{
        \begin{aligned}
            &g_{\rm{embed}} \left(\mathbf{0}\right),
            &\left| \mathbf{N}^i(\theta_n) \right| = 0; \\
            &g_{\rm{embed}} \left(\mathbf{f}^i_{\rm{meta}}\left(\theta_n\right)\right),
            &\textrm{Otherwise}.\\
        \end{aligned}
    \right.
\end{equation}
Here, $g_{\rm{embed}}$ is the embedding function that contains 2 fully connected layers with 64 output units.
${\rm{ReLU}}$ activation is used in the first layer while $\tanh$ is used in the second layer.

\MODEL~represents the \emph{\textbf{Spatial}} interactive context at the current step ($t = t_h$) through a serialized form.
A natural thought is to handle this sequence $\mathbf{f}^i \in \mathbb{R}^{N_\theta \times d_{\rm{sc}}}$ along with the observed trajectory $\mathbf{X}^i \in \mathbb{R}^{t_h \times 2}$ to learn the attentive portions inner these sequences simultaneously.
In detail, we treat the spatial \MODEL~$\mathbf{f}^i$ as a \textbf{Virtual \emph{Temporal} Sequence} that shares the same sequence length as the trajectory $\mathbf{X}^i$ or its representations.
As shown in \FIG{fig_method_backbone}, $\mathbf{f}^i$ will be padded at first (we set $N_\theta \leq t_h$).
Formally,
\begin{equation}
    \label{eq_padding}
    \mathbf{f}^i_{\rm{pad}} = \left(\vphantom{\mathbf{f}^i \left(\theta_{N_\theta}\right)}\right.
        \underbrace{
            \mathbf{f}^i \left(\theta_1\right),
            ...,
            \mathbf{f}^i \left(\theta_{N_\theta}\right)
        }_{N_{\theta}},
        \underbrace{\vphantom{\mathbf{f}^i \left(\theta_{N_\theta}\right)}
            \mathbf{0},
            ...,
            \mathbf{0}
        }_{t_h - N_{\theta}}
    \left.\vphantom{\mathbf{f}^i \left(\theta_{N_\theta}\right)}\right)^\top
    \in \mathbb{R}^{t_h \times d_{\rm{sc}}}.
\end{equation}

In most previous works \cite{alahi2016social,gupta2018social}, agent-$i$'s observed trajectory $\mathbf{X}^i$ will be first embedded into the high-dimensional $\mathbf{f}^i_{\rm{traj}} \in \mathbb{R}^{t_h \times d}$ by some embedding layer $f_{\rm{embed}}$, \IE, $
    \mathbf{f}^i_{\rm{traj}} = f_{\rm{embed}}\left(\mathbf{X}^i\right)
$.
Denote the computation of one trajectory prediction model (we call the \emph{backbone prediction model}) as $B_{\rm{pred}}$, future trajectories $\hat{\mathbf{Y}}^i$ are usually predicted by
\begin{align}
    \hat{\mathbf{Y}}^i = B_{\rm{pred}} \left(
        \mathbf{f}^i_{\rm{traj}},
        \mathbf{f}^i_{\rm{social}},
        \mathbf{f}^i_{\rm{others}}
    \right),
\end{align}
where $\mathbf{f}^i_{\rm{social}}$ denotes the original social representations in the backbone prediction model, and $\mathbf{f}^i_{\rm{others}}$ denotes other required features (like visual features from scene image $\mathbf{I}_{t_h}$).

\MODEL~Models (the \emph{\MODEL-lized} backbone prediction models) take the fused vector $\mathbf{f}^i_{\rm{fuse}}$ containing both trajectory information and interactive context to instead the single $\mathbf{f}^i_{\rm{traj}}$ to learn the temporal-attentive portions in the observed trajectories and the angle-attentive portions in the \MODEL~simultaneously.
The $\mathbf{f}^i_{\rm{fuse}}$ is fused by
\begin{equation}
    \mathbf{f}^i_{\rm{fuse}} = \tanh \left(
        \mathbf{W}_{\rm{fuse}}~{\rm{Concat}} \left(
            \mathbf{f}^i_{\rm{traj}}, 
            \mathbf{f}^i_{\rm{pad}}
        \right) + \mathbf{b}_{\rm{fuse}}
    \right).
\end{equation}
Here, $\mathbf{W}_{\rm{fuse}}$ and $\mathbf{b}_{\rm{fuse}}$ are the trainable weights and bias.
Meanwhile, the original $\mathbf{f}^i_{\rm{social}}$ will be removed.
The trajectory prediction pipeline of \MODEL~models has become
\begin{equation}
    \hat{\mathbf{Y}}^i_{\rm{SC}} = B_{\rm{pred}} \left(
        \mathbf{f}^i_{\rm{fuse}},
        \mathbf{f}^i_{\rm{others}}
    \right).
\end{equation}

\textbf{Training.}
\MODEL~does not introduce additional new loss functions.
We take Transformer\cite{vaswani2017attention}, \MSN\MSNCITE, \VMODEL\VCITE, \EVMODEL\EVCITE~as backbone trajectory prediction models to validate \MODEL's performance in our experiments.
These models will be trained with original loss functions and settings reported in their papers.

\section{Experiments}

\begin{table*}[htb]
    \footnotesize
    \centering
    \begin{tabular}{c|ccccc|c|}
        \toprule
        Models (ETH-UCY)
        & eth
        & hotel
        & univ  
        & zara1  
        & zara2  
        & ETH-UCY \\
    
        \midrule
        \SHENET\SHENETCITE~(2022) & 0.41/0.61 & 0.13/0.20 & 0.25/0.43 & 0.21/0.32 & 0.15/0.26 & 0.23/0.36 \\
        \MID\MIDCITE~(2022) & 0.39/0.66 & 0.13/0.22 & \MARK{0.22}/0.45 & \MARK{0.17}/0.30 & \MARK{0.13}/0.27 & 0.21/0.38 \\
        \EQMOTION\EQMOTIONCITE~(2023) & 0.40/0.61 & \MARK{0.12}/0.18 & 0.23/0.43 & 0.18/0.32 & \MARK{0.13}/\MARK{0.23} & 0.21/0.35 \\
        \MSN\MSNCITE~(2023) & 0.27/0.41 & \MARK{0.11}/0.17 & 0.28/0.48 & 0.22/0.36 & 0.18/0.29 & 0.21/0.34 \\
        \LED\LEDCITE~(2023) & 0.39/0.58 & \MARK{0.11}/0.17 & 0.26/0.43 & 0.18/0.26 & \MARK{0.13}/\MARK{0.22} & 0.21/0.33 \\
        \TRAJECTRONPP\TRAJECTRONPPCITE~(2020) & 0.43/0.86 & \MARK{0.12}/0.19 & \MARK{0.22}/0.43 & \MARK{0.17}/0.32 & \MARK{0.12}/0.25 & 0.20/0.39 \\
        \AGENTFORMER\AGENTFORMERCITE~(2021) & 0.26/0.39 & \MARK{0.11}/\MARK{0.14} & 0.26/0.46 & \MARK{0.15}/\MARK{0.23} & \MARK{0.14}/\MARK{0.23} & \MARK{0.18}/\MARK{0.29} \\
        \VMODEL\VCITE~(2022) & \MARK{0.23}/0.37 & \MARK{0.10}/\MARK{0.16} & \MARK{0.21}/\MARK{0.35} & 0.19/0.30 & \MARK{0.14}/\MARK{0.24} & \MARK{0.18}/\MARK{0.28} \\
        \YNET\YNETCITE~(2021) & 0.28/\MARK{0.33} & \MARK{0.10}/\MARK{0.14} & 0.24/0.41 & \MARK{0.17}/0.27 & \MARK{0.13}/\MARK{0.22} & \MARK{0.18}/\MARK{0.27} \\
        \EVMODEL\EVCITE~(2023) & \MARK{0.25}/0.38 & \MARK{0.11}/\MARK{0.16} & \MARK{0.20}/\MARK{0.34} & 0.19/0.30 & \MARK{0.13}/\MARK{0.24} & \MARK{0.17}/\MARK{0.28} \\
        
        \midrule
        \MSN-SC (Ours) & 0.27/0.39 & 0.13/0.18 & \MARK{0.22}/0.45 & 0.18/0.34 & 0.15/0.27 & \MARK{0.19}/0.33 \\
        \VMODEL-SC (Ours) & \MARK{0.25}/0.37 & \MARK{0.12}/\MARK{0.15} & \MARK{0.21}/\MARK{0.35} & \MARK{0.17}/0.29 & \MARK{0.13}/\MARK{0.22} & \MARK{0.17}/\MARK{0.27} \\
        \EVMODEL-SC (Ours) & \MARK{0.25}/0.38 & \MARK{0.12}/\MARK{0.14} & \MARK{0.20}/\MARK{0.34} & 0.18/0.29 & \MARK{0.13}/\MARK{0.22} & \MARK{0.17}/\MARK{0.27} \\
    
        \bottomrule
    \end{tabular}
    \begin{tabular}{|c|c}
        \toprule
        Models (SDD) & SDD \\

        \midrule
        \FLOWCHAIN\FLOWCHAINCITE~(2023) & 9.93/17.17 \\
        \SHENET\SHENETCITE~(2022) & 9.01/13.24 \\
        \IMP\IMPCITE~(2023) & 8.98/15.54 \\
        \LED\LEDCITE~(2023) & 8.48/11.36 \\
        \MID\MIDCITE~(2022) & 7.91/14.50 \\
        \YNET\YNETCITE~(2021) & 7.85/11.85 \\
        \MSN\MSNCITE~(2023) & 7.69/12.16 \\
        \VMODEL\VCITE~(2022) & 7.12/11.39 \\
        \EVMODEL\EVCITE~(2023) & \MARK{6.57}/\MARK{10.49} \\
        \NSPSFM\NSPSFMCITE~(2022) & \MARK{6.52}/\MARK{10.61} \\
        
        \midrule
        \MSN-SC (Ours) & 7.49/12.12 \\
        \VMODEL-SC (Ours) & \MARK{6.71}/\MARK{10.66} \\
        \EVMODEL-SC (Ours) & \MARK{6.54}/\MARK{10.36} \\
        
        \bottomrule
    \end{tabular}
    \caption{
        Comparisons on ETH-UCY (left) and SDD (right) under \emph{best-of-20} and with $t_h = 8$ frames' (3.2s) observations to predict future $t_f = 12$ frames' (4.8s) trajectories.
        Metrics are reported as ``ADE/FDE''.
        Lower ADE and FDE indicate better prediction performance.
    }
    \label{tab_ethucy}
  \end{table*}


\HARDINDENT\textbf{Datasets.}\footnote{
    \textbf{Ethics Note}: Datasets used in this work are publicly available and do not contain any sensitive personally identifiable information.
}
(a) \textbf{ETH}\cite{pellegrini2009youll}-\textbf{UCY}\cite{lerner2007crowds} contains several videos captured in pedestrian walking scenes.
We use the \emph{leave-one-out} strategy \cite{alahi2016social} to train with $\left\{t_h = 8, t_f = 12\right\}$ and sample interval $\Delta t = 0.4$s.
(b) \textbf{Stanford Drone Dataset}~\cite{robicquet2016learning} (SDD) has 60 drone videos.
Different categories of agents are annotated in pixels.
Following \cite{liang2020simaug}, we split 60\% videos to train, 20\% to validate, and 20\% to test under $\left(t_h, t_f, \Delta t\right) = (8, 12, 0.4)$.
(c) \textbf{NBA SportVU}~\cite{linou2016nba} (NBA) includes trajectories captured by the SportVU tracking system in NBA games.
Following \cite{xu2022groupnet,xu2022remember}, we set $\left(t_h, t_f, \Delta t\right) = (5, 10, 0.4)$, and sample 50K trajectories, including 65\% to train, 25\% to test, and 10\% to validate.

\textbf{Metrics.}
We measure prediction accuracy with the best Average/Final Displacement Error over 20 generated trajectories (\IE, minADE$_{20}$/minFDE$_{20}$ \cite{alahi2016social,gupta2018social}, short for ADE/FDE).
See their detailed definitions in the Appendix.

\textbf{Implementation details.}
Models are trained on one NVIDIA Tesla T4 GPU.
\MODEL~is computed on each agent's 50 nearest neighbors to save the computation resource.
For \MODEL~models, we set $N_\theta = t_h$\footnote{
    See detailed analyses about $N_\theta$ in the Appendix.
}, and set $\theta_n = 2n\pi/N_\theta \left(n = 1, 2, ..., N_\theta\right)$.
Feature dimensions $d$ and $d_{\rm{sc}}$ are set to 64.
Following \cite{zhang2020social}, trajectories are pre-processed by moving to $(0, 0)$.
We set the learning rate to 1e-4, epochs to 600, and batch size to 1500.

\begin{table}[t]
    \footnotesize
    \centering
    \begin{tabular}{c|ccc}
        \toprule
        Models (NBA) & Metrics@2.0s & Metrics@4.0s \\

        \midrule
        \PECNET\PECNETCITE~(2020) & 0.96/1.69 & 1.83/ 3.41 \\
        NMMP\cite{hu2020collaborative}~(2020) & 0.70/1.11 & 1.33/ 2.05 \\
        \VMODEL\VCITE~(2022) & 0.69/0.96 & 1.28/ 1.68 \\
        \EVMODEL\EVCITE~(2023) & 0.68/ \MARK{0.93} & 1.26/ 1.64 \\
        MemoNet\cite{xu2022remember}~(2022) & 0.71/1.14 & 1.25/ \MARK{1.47} \\
        GroupNet+NMMP\cite{xu2022groupnet}~(2022) & 0.69/1.08 & 1.25/ 1.80 \\
        GroupNet+CVAE\cite{xu2022groupnet}~(2022) & \MARK{0.62}/\MARK{0.95} & \MARK{1.13}/1.69 \\

        \midrule
        \VMODEL-SC (Ours) & \MARK{0.67}/\MARK{0.92} & 1.22/\MARK{1.51} \\
        \EVMODEL-SC (Ours) & \MARK{0.67}/\MARK{0.90} & \MARK{1.18}/\MARK{1.46} \\
        
        \bottomrule
    \end{tabular}
    \caption{
        Comparisons on NBA under \emph{best-of-20} in meters ($t_h = 5, t_f = 10$).
        Metrics are reported as ``ADE/FDE'' at different prediction lengths, including 2.0s (5 frames) and 4.0s (10 frames).
    }
    \label{tab_nba}
\end{table}

\subsection{Comparisons to State-of-the-Art Methods}

\HARDINDENT\textbf{ETH-UCY.}
\TABLE{tab_ethucy} illustrates that \MODEL~models are competitive.
In detail, the \EVMODEL-SC has achieved a noteworthy prediction performance with 5.6\% better ADE and 6.9\% better FDE compared with \AGENTFORMER.
Moreover, \MSN-SC performs better than \EQMOTION~with the improvement of 9.5\% ADE and 5.7\% FDE, even though \MSN~performs not as well as other new approaches.

\textbf{SDD.}
In \TABLE{tab_ethucy}, \VMODEL-SC outperforms \YNET~by 14.52\% ADE and 10.04\% FDE.
It has also obtained 20.87\% ADE and 6.16\% FDE improvements compared to the newly published \LED.
In addition, \EVMODEL-SC outperforms the state-of-the-art NSP-SFM by as much as 2.36\% FDE.

\textbf{NBA.}
In \TABLE{tab_nba}, compared with GroupNet+CVAE, \EVMODEL-SC's ADE is not as well as that model (about 4.42\% worse), but its FDEs (both at 2.0s and 4.0s) are better than those for about 5.26\% and 13.60\%.
In addition, even though the FDE (4.0s) of MemoNet and \EVMODEL-SC are at the same level, \EVMODEL-SC outperforms the other for about 5.60\% ADE (4.0s).
Although \EVMODEL~performs not as well as these newly published methods, the proposed \MODEL~makes it available to achieve competitive results.

\begin{figure}[t]
    \centering
    \includegraphics[width=1.0\linewidth]{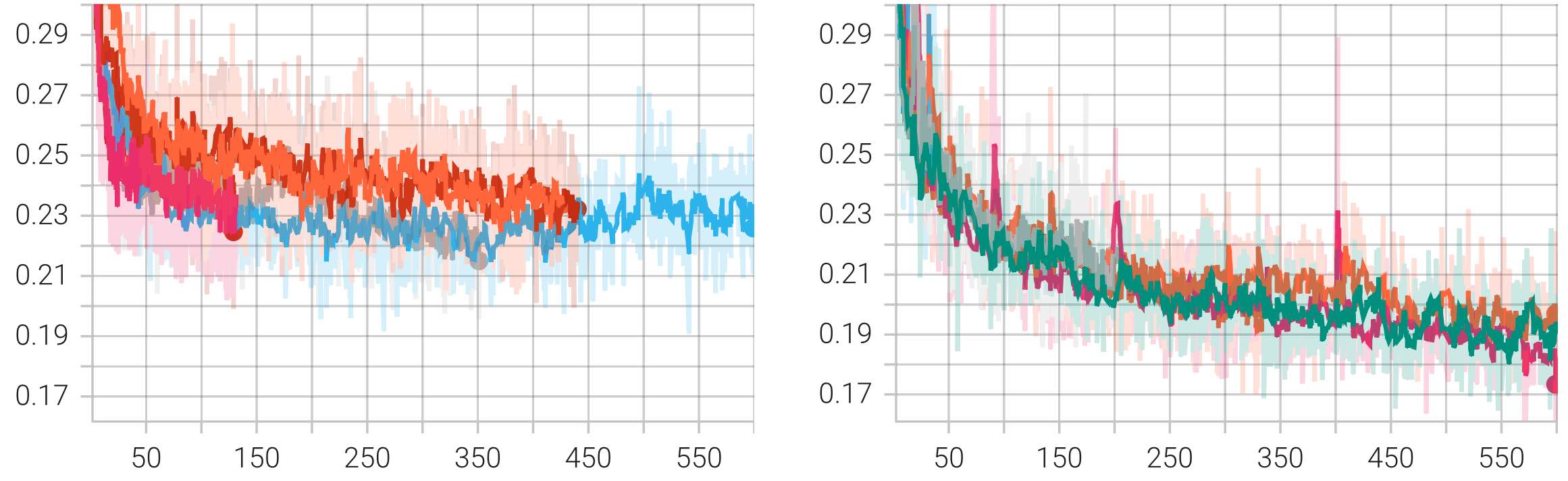}
    \caption{
        Loss curves (loss values after different training epochs) of the simple Transformer (left) and the corresponding Transformer-SC variation (right) during the training process on SDD with 600 epochs in total.
        Loss values are normalized, and each figure includes six training runs (``nan'' loss values are not displayed).
        Curves are smoothed with the decay factor $= 0.8$.
    }
    \label{fig_loss}
\end{figure}

\begin{table}[htb]
    \footnotesize
    \centering
    \begin{tabular}{c|c|cc}
        \toprule
        Variations & V D R & ADE/FDE & ADE/FDE Gain (\%) \\

        \midrule
        Transformer* & \xm \xm \xm & 17.44/33.36 & -5.89\%/-4.00\% \\
        Transformer-SC & \cm \cm \cm & 16.47/32.08 & (base) \\

        \midrule
        \MSN* & \xm  \xm  \xm & 7.79/13.09 & -4.01\%/-8.00\%  \\
        \MSN-SC-a1 & \xm \cm \cm & 7.53/12.30 & -0.53\%/-1.49\% \\
        \MSN-SC-a2 & \cm \xm \cm & 7.57/12.40 & -1.07\%/-2.31\%\\
        \MSN-SC-a3 & \cm \cm \xm & 7.60/12.52 & -1.47\%/-3.30\% \\
        \MSN-SC & \cm \cm \cm & 7.49/12.12 & (base) \\

        \midrule
        \VMODEL* & \xm \xm \xm & 7.04/10.94 & -4.92\%/-2.63\% \\
        \VMODEL-SC-a1 & \xm \cm \cm & 6.86/10.82 & -2.24\%/-1.50\% \\
        \VMODEL-SC-a2 & \cm\ \xm \cm & 6.87/10.87 & -2.38\%/-1.97\% \\
        \VMODEL-SC-a3 & \cm \cm \xm & 6.78/10.71 & -1.04\%/-0.47\% \\
        \VMODEL-SC & \cm \cm \cm & 6.71/10.66 & (base) \\

        \midrule
        \EVMODEL* & \xm \xm \xm & 6.73/10.75 & -2.91\%/-3.76\% \\
        \EVMODEL-SC-a1 & \xm \cm \cm & 6.67/10.73 & -1.99\%/-3.57\% \\
        \EVMODEL-SC-a2 & \cm \xm \cm & 6.64/10.55 & -1.53\%/-1.83\% \\
        \EVMODEL-SC-a3 & \cm \cm \xm & 6.59/10.48 & -0.76\%/-1.16\% \\
        \EVMODEL-SC & \cm \cm \cm & 6.54/10.36 & (base) \\
        
        \bottomrule
    \end{tabular}
    \caption{
        Ablation studies on validating \MODEL~meta components on SDD.
        ``V'', ``D'', and ``R'' indicate whether $\mathbf{f}^i_{\rm{vel}}$, $\mathbf{f}^i_{\rm{dis}}$, or $\mathbf{f}^i_{\rm{dir}}$ are included in the meta vector $\mathbf{f}^i_{\rm{meta}}$.
        Models with ``*'' are reproduced under the same condition.
    }
    \label{tab_ab_factors}
\end{table}

\begin{figure*}[htb]
    \centering
    \includegraphics[width=0.95\linewidth]{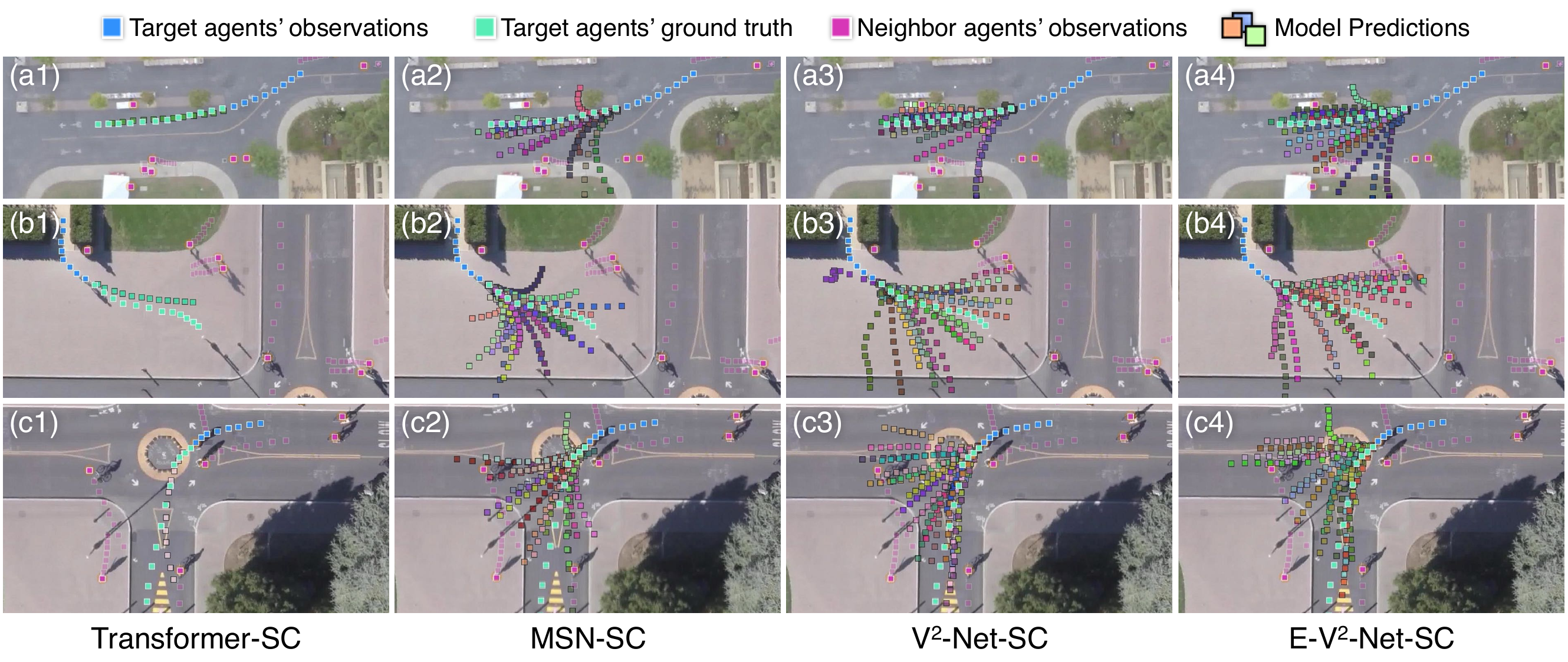}
    \caption{
        Visualized predictions of \MODEL~models with different backbone prediction models in several ETH-UCY and SDD scenes.
    }
    \label{fig_vis}
\end{figure*}

\begin{figure*}[htb]
    \centering
    \includegraphics[width=0.95\linewidth]{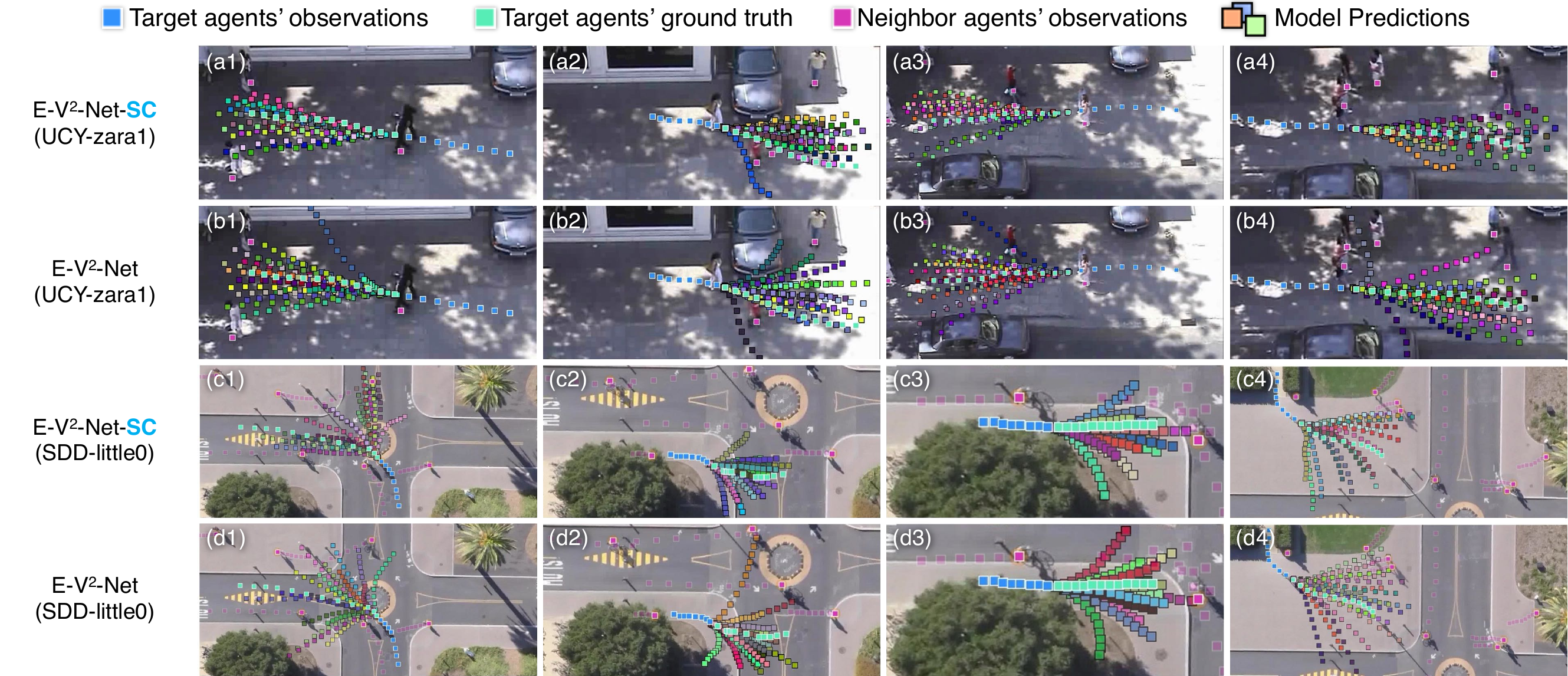}
    \caption{
        Prediction comparisons of \MODEL~models and their original models in several interactive ETH-UCY and SDD scenes.
    }
    \label{fig_social}
\end{figure*}

\subsection{Ablation Studies and Quantitative Analyses}

\HARDINDENT\textbf{Overall Validation of \MODEL.}
As shown in \TABLE{tab_ab_factors}, \MODEL~could help even the simplest Transformer \cite{vaswani2017attention} (which considers nothing about agents' multimodality and interactions) improve 5.89\% ADE and 4.00\% FDE.
\MODEL~also facilitates \MSN, \VMODEL, and \EVMODEL~with up to 4.92\% ADE gains and 8.00\% FDE gains compared to their original models.
In addition, we also plot loss curves ($\ell_2$ loss) of 6 training runs of the simplest Transformer and the Transformer-SC in \FIG{fig_loss}.
It shows that \MODEL~could help the loss drop faster for Transformer-SC with an average of 13.04\% lower than that in the Transformer after 600 training epochs.
Moreover, 5 out of 6 Transformer runs' loss values fall into ``nan'' due to the inappropriate gradients before 450 training epochs.
As for Transformer-SC, only two runs are terminated, demonstrating that \MODEL~may also somehow act as a normalization factor, thus improving the training stability.

\textbf{Validation of \MODEL~Meta Components.}
As shown in \FIG{tab_ab_factors}, both \VMODEL~and \EVMODEL~benefit further from the velocity and distance factors, each of which could provide at least 1.50\% ADE and FDE gains (SC-a1 and SC-a2 variations).
However, \MSN-SC variations are not so sensitive to these two factors (less than 1.07\% ADE differences among \MSN-SC, \MSN-SC-a1, and \MSN-SC-a2) but rely more on the direction factor (up to 3.30\% FDE gain has been brought by this factor).
Although these factors contribute differently to different backbone models, the combination of all these factors promotes the most, for removing each one could lead to a serious performance drop.

\textbf{Parameters \& Efficiency.}
Please refer to the Appendix.

\begin{figure}[htb]
    \centering
    \includegraphics[width=1.0\linewidth]{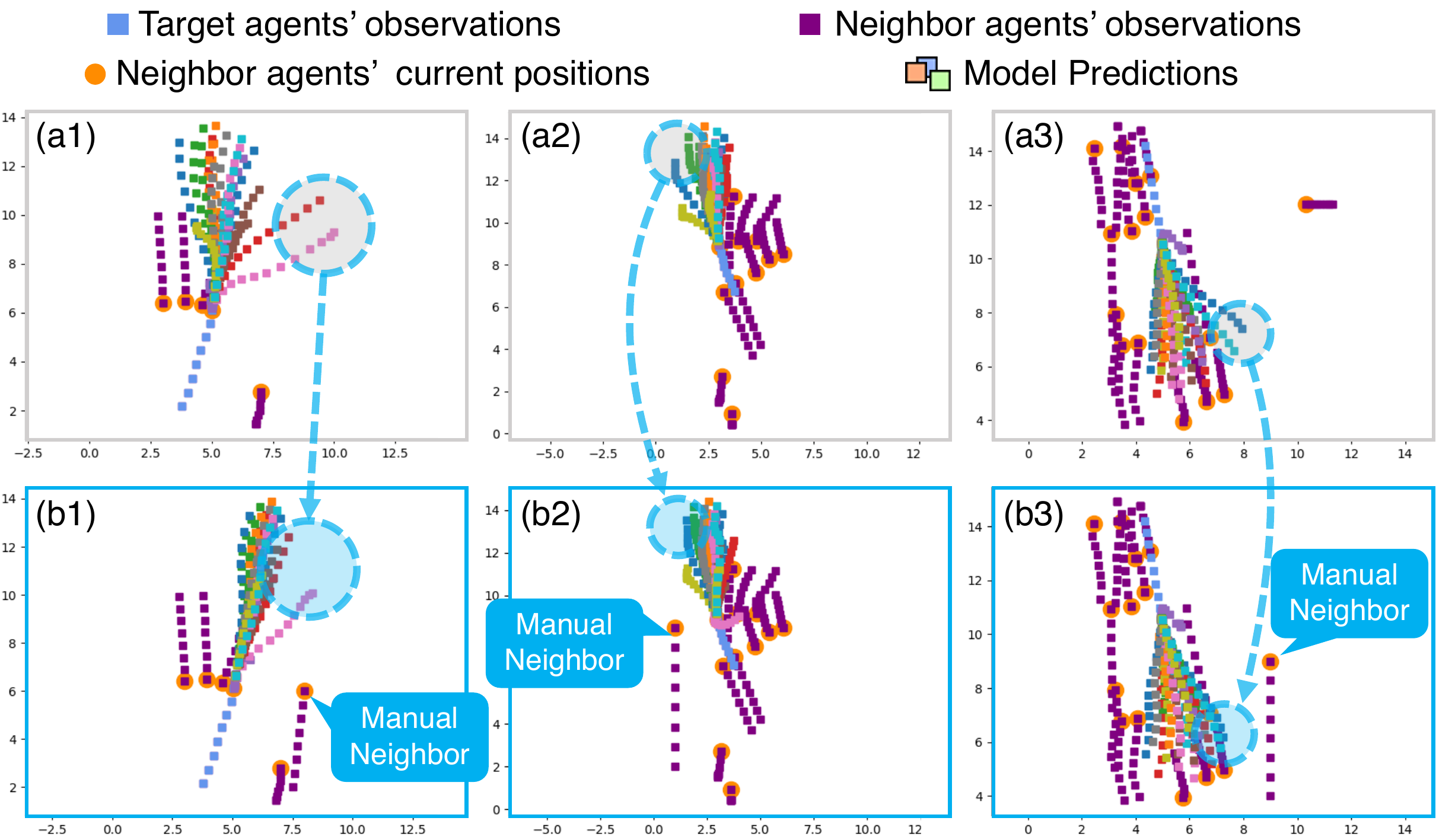}
    \caption{
        Toy Examples I: Validation of social interactions.
        The arrows and circles point to the same predicted trajectory that has been changed significantly due to the manual neighbor.
    }
    \label{fig_toy}
\end{figure}

\begin{figure}[htb]
    \centering
    \includegraphics[width=1.0\linewidth]{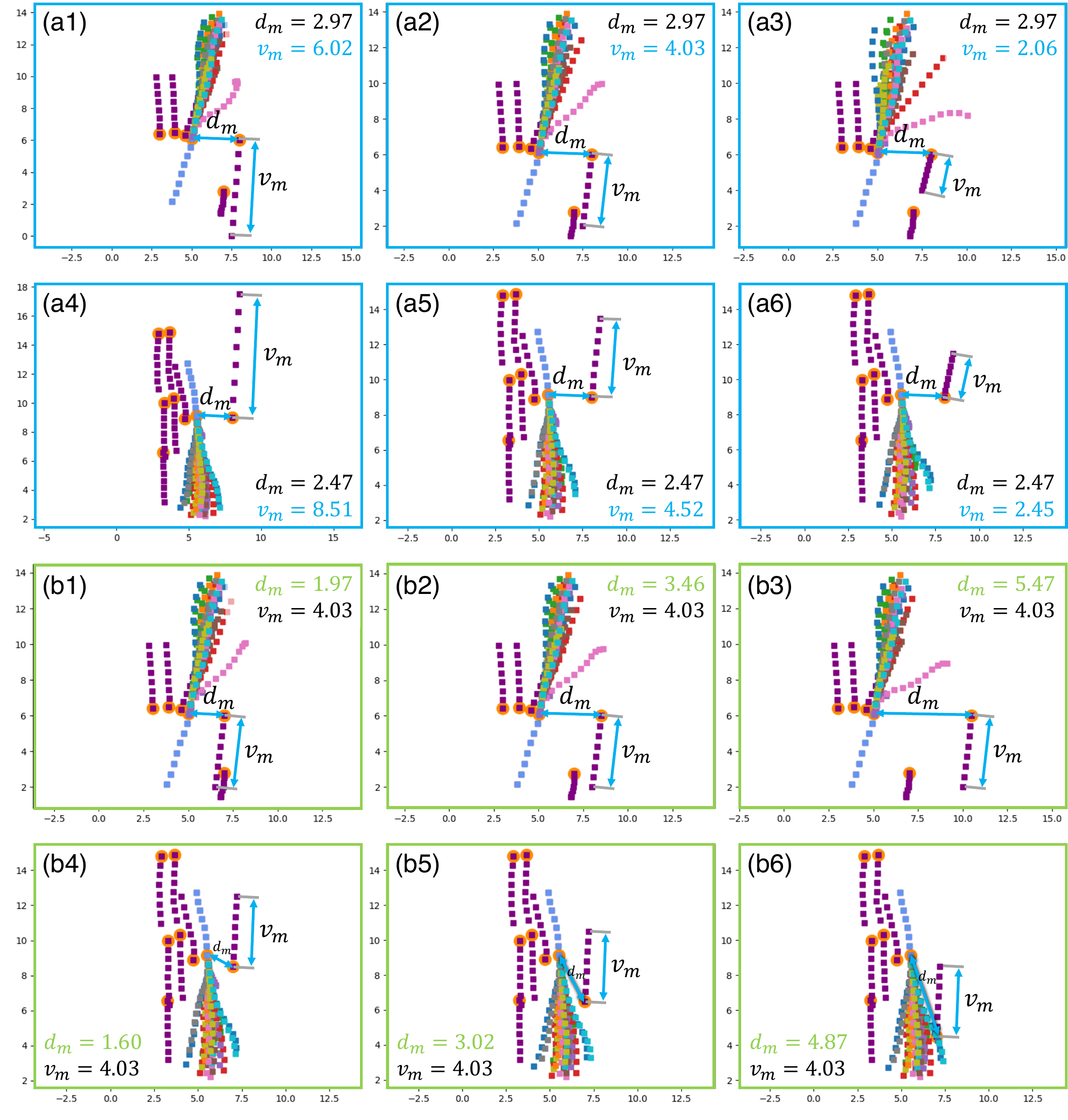}
    \caption{
        Toy Examples II: Validation of \MODEL~meta components by manually changing the manual neighbor's velocity (denoted by $v_m$) and its distance to the target agent ($d_m$).
    }
    \label{fig_social_ab}
\end{figure}

\begin{figure}[htb]
    \centering
    \includegraphics[width=1.0\linewidth]{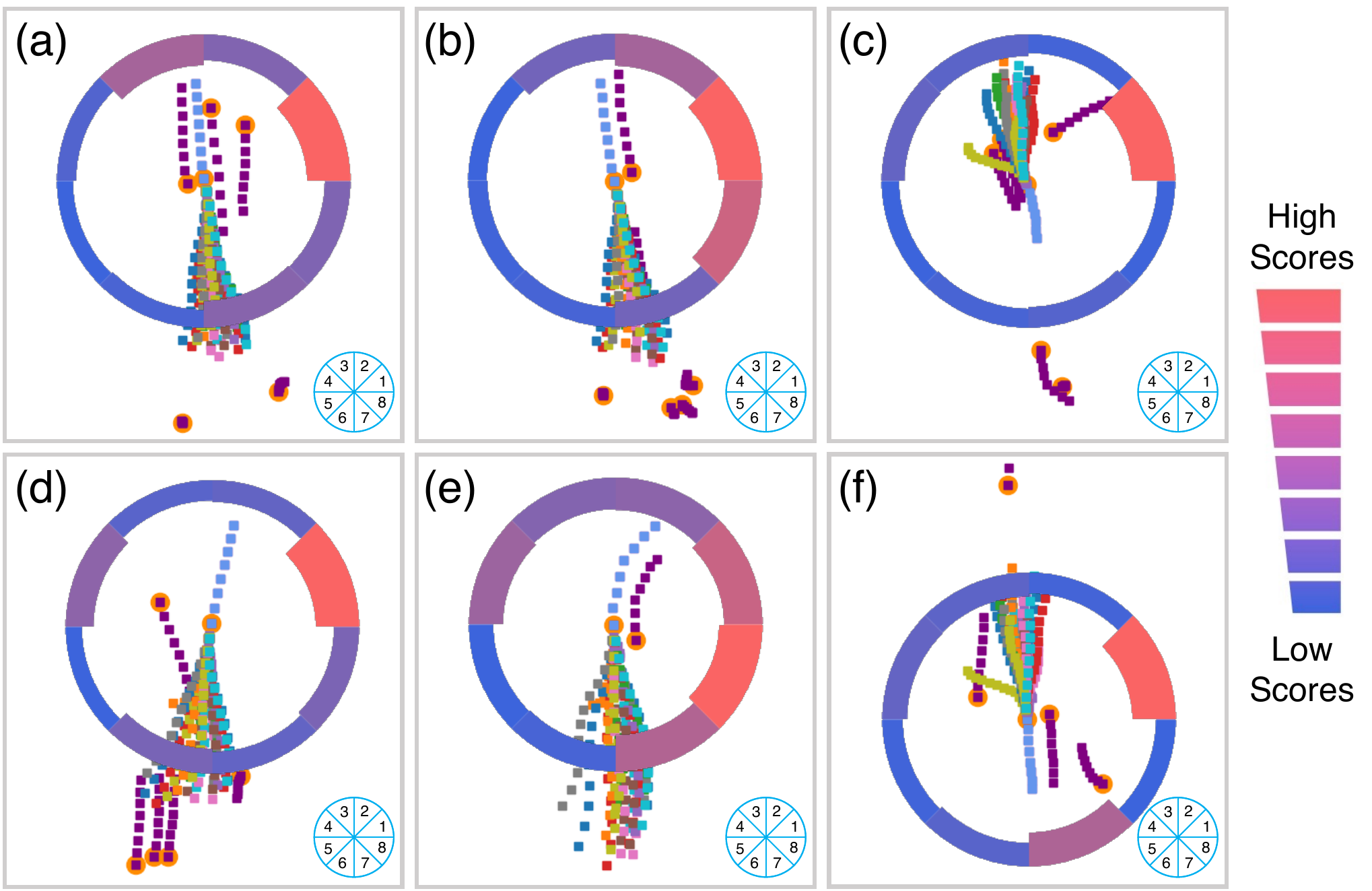}
    \caption{
        Visualized attention scores of each \MODEL~partition in the colored-ring form on several ETH-UCY prediction scenes.
        Partitions with higher attention scores (redder and wider) contribute more to the final predicted trajectories.
    }
    \label{fig_attention}
\end{figure}

\begin{figure}[htb]
    \centering
    \includegraphics[width=1.0\linewidth]{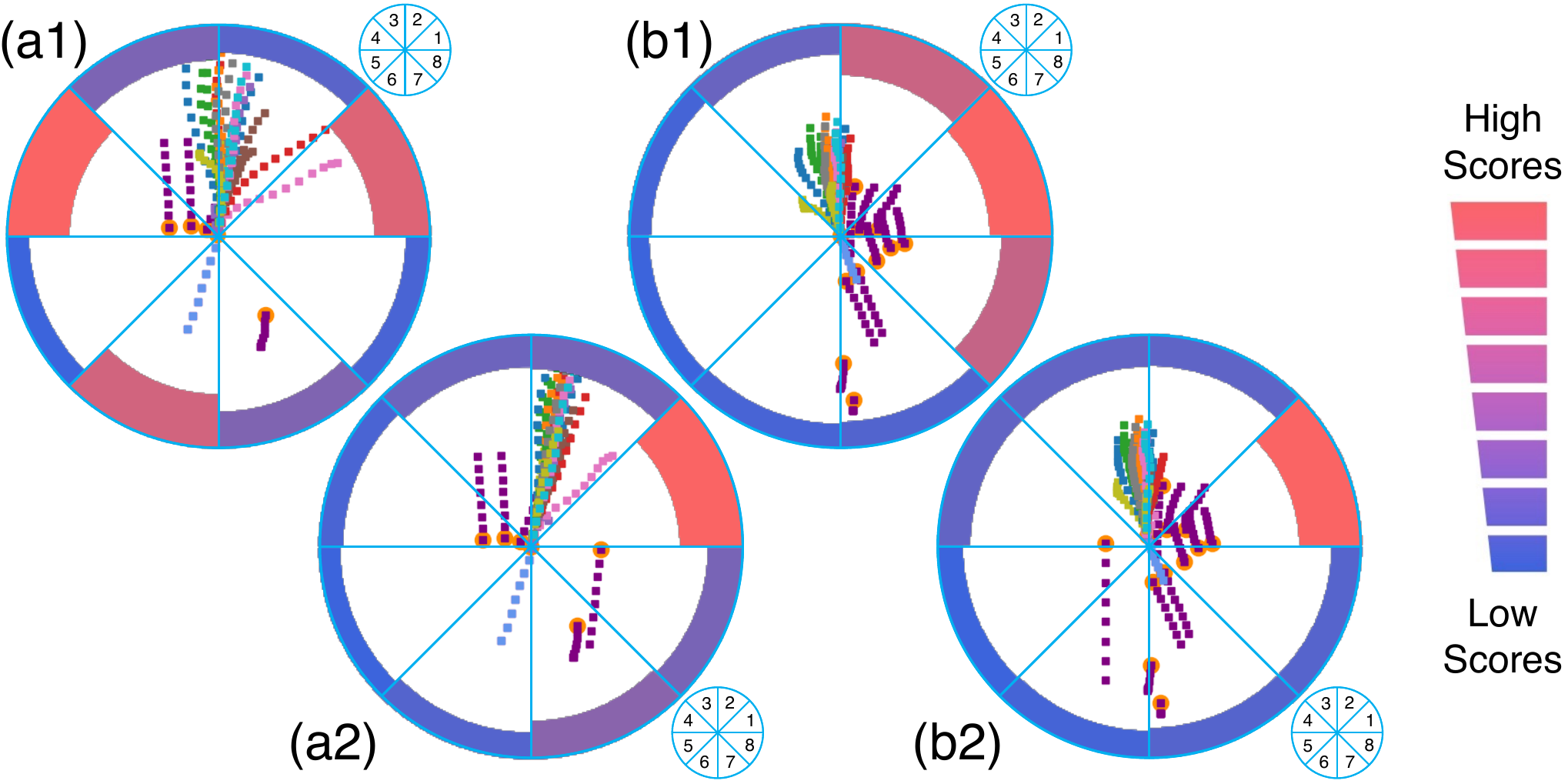}
    \caption{
        Toy Examples III: Comparisons of attention scores before and after adding manual neighbors in ETH-UCY scenes.
    }
    \label{fig_attention_toy}
\end{figure}

\subsection{Qualitative Analyses}

\HARDINDENT\textbf{Visualized Predictions.}
\FIG{fig_vis} visualizes trajectories predicted by several \MODEL~models.
We can see that all \MODEL~models' predictions present similar ways to handle social interactions.
For example, predictions in \FIG{fig_vis} (a2) to (a4) all present avoidance of the group of pedestrians standing still at the roadside.
Similarly, avoiding the upcoming biker about to cross the intersection has also become a major concern in (c1) to (c4).
These results indicate the effectiveness of the proposed \MODEL~that works with different backbone prediction models.

\textbf{Visualized Social Interaction Cases.}
As shown in \FIG{fig_social}, we observe that predictions given by \MODEL~models seem to be more ``conservative'' with a tendency to avoid others in different social interaction situations, which looks like it is trying to avoid possible collisions or too-close social distances.
For example, predictions in (a2) display heavier avoidance of the man coming from the car side than in (b2).
Similarly, the potential collision between the moving biker and the focused walking pedestrian has been caught in (c3), thus representing a prediction shift (comparing the \textbf{\color[HTML]{3187D7}{blue}} prediction in (c3) and the \textbf{\color[HTML]{B61C43}{red}} one in (d3)).
In addition, two pedestrians walk towards right together in cases (a4) and (b4).
Unlike case (b4), \MODEL~model forecasts that the target agent may not crossover the other two's potential trajectories, like it ``thought'' that the agent would rather keep its relative position in the group.
Note that social interactions are not only limited to these collision cases.
They are common behavior patterns among agents, including scene-specific interactions like the game-related interactions in the NBA dataset.
Please see the qualitative analyses of NBA interactions in the Appendix.

\textbf{Toy Examples I (Social Interactions).}
We design a series of toy examples on several real-world prediction scenes from ETH-UCY to verify the capacity of \MODEL~in handling social interactions in \FIG{fig_toy}.
We manually add new ``manual neighbors'' in different positions (partitions) to test \MODEL~model's response.
Each manual neighbor's observed trajectory is simulated by linearly interpolating from the given start and end points.
Comparing \FIG{fig_toy} (a1) and (b1), several predicted trajectories to the right of the target agent have changed a lot.
For example, the \textbf{\color[HTML]{D02324}{red trajectory}} contracts violently to the other side, which seems quite likely to prevent possible collisions with the manual neighbor that walks towards itself.
Similar phenomenons also appear in \FIG{fig_toy} (b2), whose \textbf{\color[HTML]{1D6CAB}{blue}} and \textbf{\color[HTML]{259425}{green}} predictions show varying degrees of avoidance tendencies, depending on where they are located.
The predicted trajectories colored in \textbf{\color[HTML]{1D6CAB}{blue}} and \textbf{\color[HTML]{17B6C8}{cyan}} in \FIG{fig_toy} (a3) also show a tendency to avoid the manual neighbor.
These cases illustrate the adaptivity of the \MODEL~in customizing different predictions.
We have provided more toy cases in the Appendix.

\textbf{Toy Examples II (\MODEL~Meta Components).}
\FIG{fig_social_ab} further shows the qualitative effect on predicted trajectories caused by the other two \MODEL~factors, velocity and distance.
We first focus on the velocity factor validated in (a1) to (a3) or (a4) to (a6).
When the manual neighbor has a higher velocity, the predicted trajectories will be affected by it to a greater extent, and vice versa, especially for the \textbf{{\color[HTML]{DF6CBA}{pink predictions}}} in (a1) to (a3).
Except for the velocity, as the distance $d_m$ increases in \FIG{fig_social_ab} (b1) to (b3) or (b4) to (b6), the influence of manual neighbors gradually becomes smaller.
In contrast, the predicted trajectory may produce a heavier shift when $d_m$ decreases step by step.
These ``social-like'' behaviors brought by different \MODEL~meta components are learned adaptively during training.
It illustrates \MODEL's capacity and explainability to represent potential social interaction cases and finally modify predicted trajectories socially.

\textbf{Toy Examples III (\MODEL~Partitions).}
In \FIG{fig_attention}, we visualize how each \MODEL~partition contributes by squaring-sum the feature $\mathbf{f}^i_{\rm{pad}}$ in each partition, which we call the \emph{Attention Score}.
Note that agent $i$ itself will be treated as its self-neighbor located in partition 1 (see \EQUA{eq_selfneighbor}).
We can see from these figures that \MODEL~acts far from simply locating all neighbors but tells which partitions should be paid more attention to when forecasting trajectories.
In \FIG{fig_attention} (a) to (d), partition 1 catches more attention, and others present different trends according to different neighbor distributions.
Similar to the phenomenon shown in \FIG{fig_social_ab}, neighbors that are closer to the target agent elicit higher levels of attention (like partition 3 against partition 2 in (a)), and neighbors that move faster attract more attention (partition 4 compared to partition 6 in (d)).
Unlike these cases, partition 8 owns a higher score than partition 1 in case (e), indicating that the \MODEL~cares more about the neighbor walking along with the target agent rather than the agent itself.
It is interesting in case (e) that some partitions have also been assigned scores even though there are no neighbors.

Furthermore, \FIG{fig_attention_toy} shows how the attention scores change after adding manual neighbors.
In case (a1), \MODEL~focuses more on partition 4.
However, the scores have changed significantly in case (a2) due to the manual neighbor located between partitions 1 and 8, and partition 4 has been less focused.
Cases (b1) and (b2) also show the similar trends.
All these qualitative results illustrate \MODEL's ability to describe different interactions and the effectiveness of partitioned modeling social interactions.

\textbf{Limitations.}
\MODEL~does not contain the directions where neighbor agents came from.
It also does not directly consider the interactions among neighbor agents and how they affect the target agent, \IE, the \emph{high-order} interactions.
We will further study it in our subsequent works.

\section{Conclusion}

This work focuses on the modeling of social interactions when forecasting pedestrian trajectories.
Like marine animals localizing and communicating with others through echolocation, we bring a simple priori rule to construct the angle-based social interaction representation \MODEL, which aims to learn how three meta components modify agent trajectories.
Experiments on multiple datasets show its competitiveness, and additional toy experiments also prove its effectiveness in handling social interactions.

~\\\textbf{Acknowledgement}
This project is supported by National Natural Science Foundation of China (62172177), and National Key R\&D Program of China (2022YFC3301000).

{
    \small
    \bibliographystyle{ieeenat_fullname}
    \bibliography{ref.bib}
}


\section*{\appendixname}

\appendix

\section{Definitions of Metrics and Attention Scores}
\label{section_appendix_attention}

\noindent\textbf{Metrics.}
We evaluate prediction accuracy using the Average/Final Displacement Error (known as ADE and FDE) \cite{alahi2016social,gupta2018social}.
Models are validated by the best metrics computed from 20 randomly generated trajectories for each case (\emph{best-of-20}, \IE, minADE$_{20}$ and minFDE$_{20}$).
For agent $i$, we have
\begin{align}
    {\rm{minADE}}_{20}
    \left(
        \mathbf{Y}^i, \left\{\hat{\mathbf{Y}}^i_k\right\}
    \right)
    &= \min_k \frac{1}{t_f}
    \sum_{t = t_h + 1}^{t_h + t_f}
    \Vert
        \mathbf{p}^i_t - 
        \hat{\mathbf{p}_k}^i_t
    \Vert_2, \\
    {\rm{minFDE}}_{20}
    \left(
        \mathbf{Y}^i, \left\{\hat{\mathbf{Y}}^i_k\right\}
    \right)
    &= \min_k
    \Vert
        \mathbf{p}^i_{t_h + t_f} - 
        \hat{\mathbf{p}_k}^i_{t_h + t_f}
    \Vert_2.
\end{align}
Here, vectors with $_k$ come from the $k$-th prediction.

\textbf{Attention Scores.}
We introduce the \emph{Attention Scores} to quantitatively analyze how each \MODEL~partition relatively contributes to the final predicted trajectories.
For the target agent $i$ and the $n$-th partition, it is defined as the normalized squared sum of each $\mathbf{f}^i \left(\theta_n\right) \in \mathbb{R}^{d_{\rm{sc}}}$.
Formally,
\begin{equation}
    {\rm{AttentionScore}}(i, n) = \frac{
        {\mathbf{f}^i \left(\theta_n\right)}^{\top} \mathbf{f}^i \left(\theta_n\right)
    }{
        \sum_{m=1}^{N_\theta} {\mathbf{f}^i \left(\theta_m\right)}^{\top} \mathbf{f}^i \left(\theta_m\right)
    }.
\end{equation}

The attention score evaluates the contribution of different partitions to the subsequent prediction network at the \textbf{feature level}, meaning that a partition with more neighbors may not directly lead to a higher score.
It is obtained through the combined effect of multiple layers together during the training process, including the embedding layers $g_{\rm{embed}}$, the fuse layer $\left\{\mathbf{W}_{\rm{fuse}}, \mathbf{b}_{\rm{fuse}}\right\}$, as well as the backbone prediction model $B_{\rm{pred}}$.
Thus, we choose this item to analyze how the \MODEL~contributes to the whole prediction model only \emph{qualitatively}.

\section{Additional Experimental Analyses on NBA SportVU Dataset}

Due to the page limitations, we only report \MODEL~models' performances on ETH-UCY and SDD with both quantitative and qualitative results.
This section further validates their detailed performance in handling different social interaction cases in the \textbf{NBA SportVU Dataset} by providing more additional qualitative results.

\begin{table}[tbp]
    \small
    \centering
    \begin{tabular}{c|ccc}
        \toprule
        Models & 
        \makecell{ADE\\(4.0s)} &
        \makecell{FDE\\(@2.0s)} &
        \makecell{FDE\\(@4.0s)} \\

        \midrule
        \SOCIALLSTM\SOCIALLSTMCITE & 1.79 & 1.53 & 3.16 \\
        \SOCIALGAN\SOCIALGANCITE & 1.62 & 1.36 & 2.51 \\
        \STGCNN\STGCNNCITE & 1.59 & 0.99 & 2.37 \\
        \STAR\STARCITE & 1.26 & 1.28 & 2.04 \\
        \PECNET\PECNETCITE & 1.83 & 1.69 & 3.41 \\
        NMMP\cite{hu2020collaborative} & 1.33 & 1.11 & 2.05 \\
        GroupNet+NMMP\cite{xu2022groupnet} & 1.25 & 1.08 & 1.80 \\
        GroupNet+CVAE\cite{xu2022groupnet} & \MARK{1.13} & \MARK{0.95} & 1.69 \\
        MemoNet\cite{xu2022remember} & 1.25 & \NA & \MARK{1.47} \\
        
        \midrule
        \VMODEL*\VCITE & 1.28 & 0.96 & 1.68 \\
        \VMODEL-SC & 1.22 & \MARK{0.92} & \MARK{1.51} \\
        \EVMODEL*\EVCITE & 1.26 & \MARK{0.93} & 1.64 \\
        \EVMODEL-SC & \MARK{1.18} & \MARK{0.90} & \MARK{1.46} \\
        
        \bottomrule
    \end{tabular}
    \caption{
        Comparisons on NBA under \emph{best-of-20} in meters.
        Lower ADE and FDE indicate better prediction performance.
        Models with ``*'' are reproduced under the same training settings.
    }
    \label{tab_nba_appendix}
\end{table}

\begin{figure}[t]
    \centering
    \includegraphics[width=1.0\linewidth]{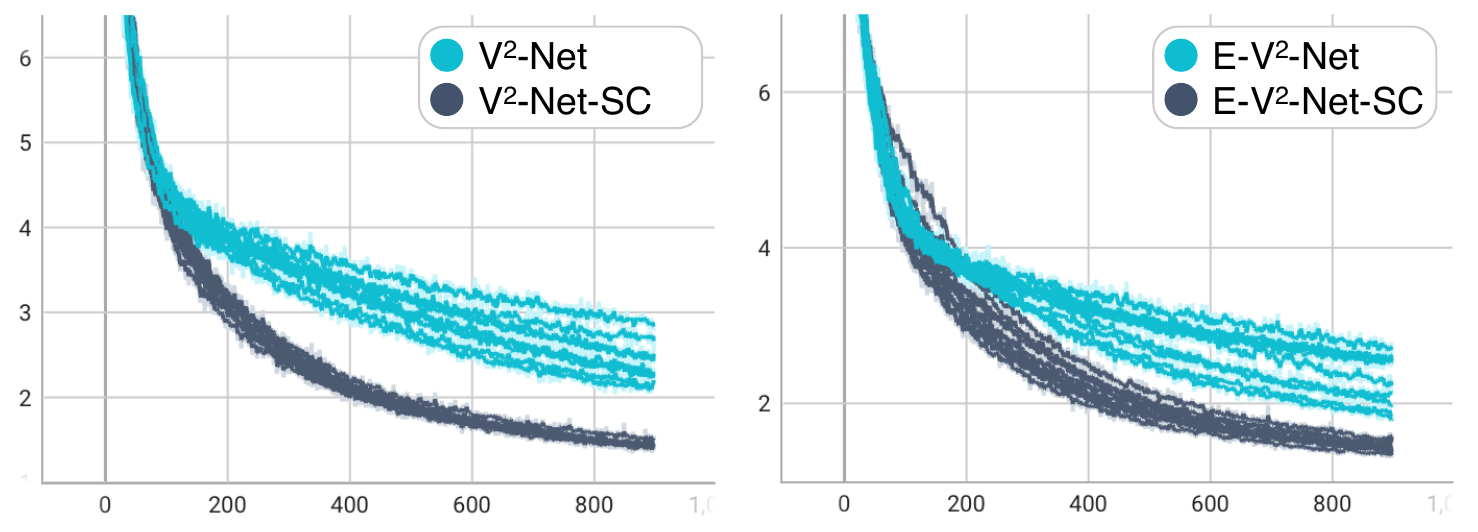}
    \caption{
        Loss curves ($\ell_2$ loss at different training epochs) of different models at different training runs on NBA dataset.
        Curves are smoothed with the decay factor $= 0.8$.
    }
    \label{fig_loss_nba}
\end{figure}

\begin{figure*}[t]
    \centering
    \includegraphics[width=1.0\linewidth]{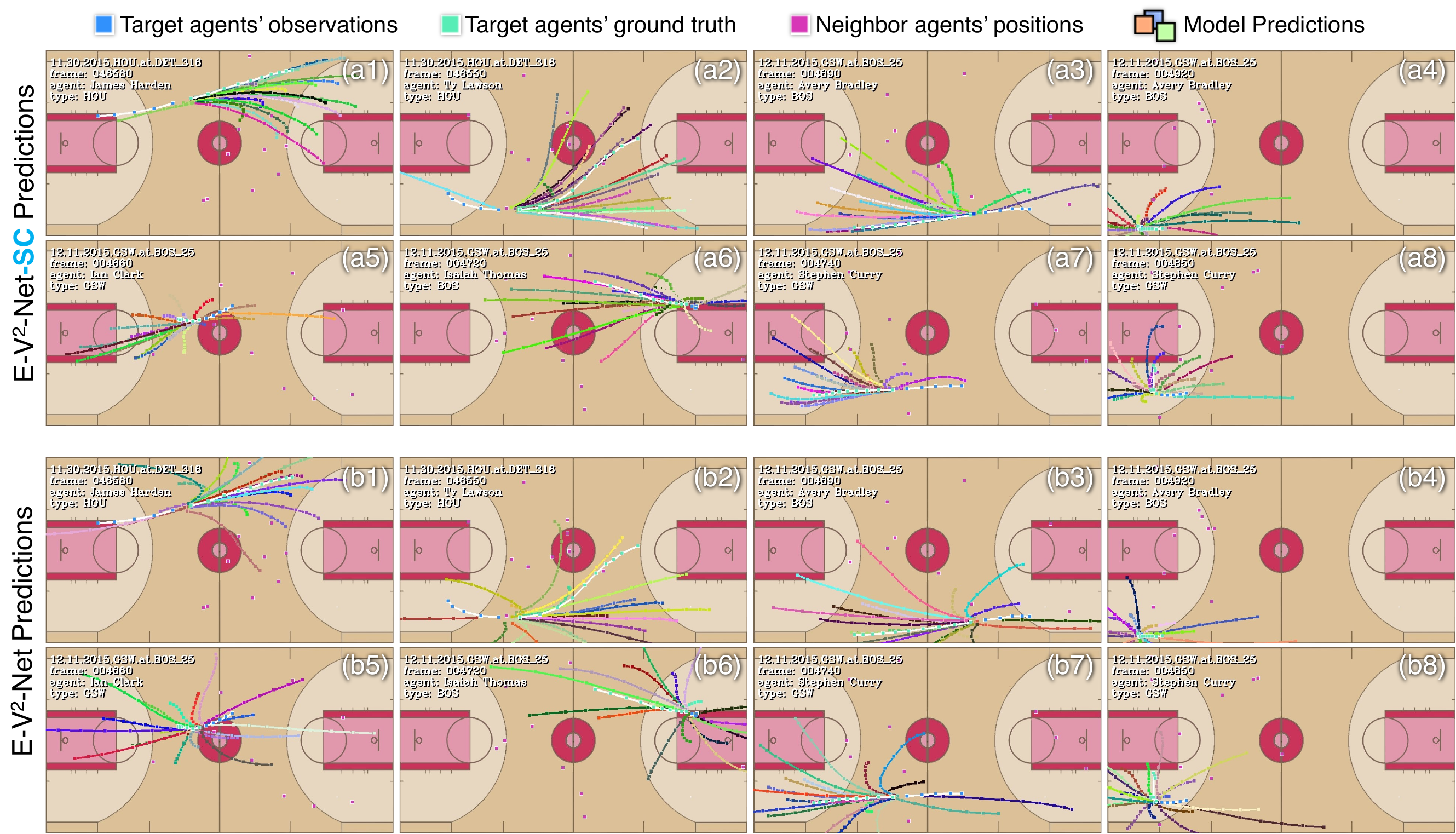}
    \caption{
        Visualized predicted trajectories provided by \MODEL~model \EVMODEL-SC (subfigures (a1) to (a8)) and the original \EVMODEL~(subfigures (b1) to (b8)) on several NBA prediction scenes.
        Each sample includes 20 randomly generated trajectories.
    }
    \label{fig_nba_vis}
\end{figure*}

\subsection{Dataset Configurations}

The \textbf{NBA SportVU Dataset} \cite{linou2016nba} (short for \textbf{NBA} dataset) is made up of a large number of real-world trajectories of ten players plus a ball captured by the SportVU tracking system during several NBA games.
The complex interactions between different players will pose significant challenges for trajectory prediction.
Positions of all players and balls are labeled in foot (1 foot = 0.3048 meter).

Following the settings of \cite{xu2022groupnet,xu2022remember}, we predict future $t_f = 10$ frames' trajectories based on the past $t_h = 5$ frames' observations.
The sample interval between two frames is still set to $\Delta t = 0.4$s.
Frames where the basketball is not on the court will be ignored.
We randomly sample about 50K prediction cases (\IE, 50K trajectories) from multiple games to validate models.
Among these cases, 65\% (about 32,500 samples) will be used for training, 25\% (about 12,500 samples) for testing, and the remaining 10\% for validation.

\subsection{Baselines}

We choose \SOCIALLSTM\SOCIALLSTMCITE,
\SOCIALGAN\SOCIALGANCITE,
\STGCNN\STGCNNCITE,
\STAR\STARCITE,
\PECNET\PECNETCITE,
NMMP\cite{hu2020collaborative},
GroupNet+NMMP\cite{xu2022groupnet},
GroupNet+CVAE\cite{xu2022groupnet},
MemoNet\cite{xu2022remember},
\VMODEL*\VCITE,
and \EVMODEL*\EVCITE~
as our baselines on NBA dataset.

\subsection{Metrics}

Except for ADE and FDE (minADE$_{20}$ and minFDE$_{20}$), following \cite{xu2022groupnet}, we use the FDE-at-$t$-moment as a new metric to measure prediction performance.
In detail, under the setting of $\left(t_h, t_f\right) = (5, 10)$ with sample interval $\Delta t = 0.4$s, the newly added metric FDE-at-5th-moment (minFDE$_{20}$@2.0s, short for FDE@2.0s) is defined as
\begin{align}
    {\rm{minFDE}_{20}}(t) &= \min_k \left\Vert
        \mathbf{p}^i_t -
        \hat{\mathbf{p}_k}^i_t
    \right\Vert_2, \\
    {\rm{FDE@2.0s}} &= {\rm{minFDE}_{20}}(t = t_h + 5).
\end{align}
The original FDE can be treated as FDE@4.0s, \IE,
\begin{equation}
    {\rm{FDE@4.0s}} = {\rm{minFDE}_{20}}(t = t_h + 10).
\end{equation}

\subsection{Quantitative Analyses}

\noindent\textbf{Comparisons to State-of-the-Art Methods.}
As shown in \TABLE{tab_nba_appendix}, the \MODEL~model \EVMODEL-SC has achieved competitive results.
Compared with the GroupNet+CVAE that obtains the best ADE, \EVMODEL-SC's ADE is not as well as that model (about 4.42\% worse ADE), but its FDEs (both at 2.0s and 4.0s) are better than those for about 5.26\% and 13.60\%.
In addition, even though the FDE@4.0s of MemoNet and \EVMODEL-SC are at the same level (less than 1\% differences), \EVMODEL-SC outperforms the other for about 5.60\% ADE.
Although the original \EVMODEL~performs not as well as these newly published methods, the proposed \MODEL~makes it available to achieve competitive results.

\textbf{Ablation Studies.}
We validate \MODEL~on two backbone models, \VMODEL~and \EVMODEL, and report their corresponding \MODEL~models' performance in \TABLE{tab_nba_appendix}.
With the help of the proposed \MODEL, both these models have achieved considerable quantitative performance gains.
In detail, compared with the basic \VMODEL, \VMODEL-SC has achieved the 4.68\% better ADE and the 10.11\% better FDE (@4.0s).
The \EVMODEL-SC also outperforms \EVMODEL~for about 6.34\% ADE and 10.97\% FDE (@4.0s).
These results indicate the quantitative effectiveness of the proposed \MODEL~for handling prediction cases with complex social interactions on NBA dataset.

\subsection{Qualitative Analyses}

\noindent\textbf{Analyses of the Training Process.}
We visualize the loss ($\ell_2$ loss) curves of \VMODEL, \EVMODEL, and their \MODEL~models at multiple training runs on NBA dataset in \FIG{fig_loss_nba}.
All these models are trained under the same settings.
It shows that the loss values drop faster and finally become lower by introducing \MODEL~to baseline models.
In addition, their loss values become more stable across different training runs compared to the original model.
We can infer that the proposed \MODEL~may also play a normalization factor, thus reducing the influence of randomized training factors (such as the shuffle operation at each training epoch and the randomly sampled noise vectors to generate multiple predictions).

\textbf{Visualizations of Social Behaviors.}
We visualize trajectories forecasted by the \MODEL~model \EVMODEL-SC and the original \EVMODEL~in several NBA scenes in \FIG{fig_nba_vis}.
These models do not take into account agents' categories (\IE, players with different teams or basketball) when forecasting trajectories.
For prediction scenes with different distributions of neighbor players, \EVMODEL-SC's predictions present better interactive trends.

Comparing \FIG{fig_nba_vis} (a1 to a4) and (b1 to b4), several trajectories predicted by the non-\MODEL~model (b1 to b4) have gone out of the court, while there are rarely these cases in the predictions of \MODEL~model (a1 to a4).
It shows that \MODEL~models could learn players' different behavior patterns according to the \MODEL, even though they do not know where the borders of the court are, thus making their predictions in line with the scene context.

In addition, the game-related interaction is a class of interactions specific to the NBA dataset, such as players carrying the ball on offense, switching from offense to defense, and many other interactive behaviors.
Comparing \FIG{fig_nba_vis} (a5 to a8) and (b5 to b8), we can see that \MODEL~could also better describe these interactive behaviors.
For example, agent ``Isaiah Thomas'' moves from a complete standstill to start moving from the free throw lane during the observation period in case (a6).
According to other players' status, the \MODEL~model finally provides predictions that seem like running to the frontcourt to start the offense.
Unlike predictions shown in \FIG{fig_nba_vis} (a6), trajectories predicted by the non-\MODEL~model appear very confusing, including both aggressive and defensive.
Other game-interactive cases, like scoring in various ways in case (a7) and the flexible movements in case (a8), present similar trends, which indicates \MODEL's capability to handle various social-interactive behaviors in different prediction scenes.

\section{Additional Experimental Analyses on nuScenes Dataset}

\MODEL~is proposed to handle interactions among pedestrians.
In this section, we conduct a series of experiments on the nuScenes dataset \cite{caesar2020nuscenes,caesar2019nuscenes} to further validate how \MODEL s~model interactions among vehicles as well as how they perform in traffic prediction scenes.

\begin{table}[tbp]
    \small
    \centering
    \begin{tabular}{c|cc|cc}
        \toprule
        Models & ADE$_{5}$ & FDE$_{5}$ & ADE$_{10}$ & FDE$_{10}$ \\

        \midrule
        \TRAJECTRONPP\TRAJECTRONPPCITE & 3.14 & 7.45 & 2.46 & 5.65 \\
        \YNET\YNETCITE & 2.46 & 5.15 & 1.88 & 3.47 \\
        \AGENTFORMER\AGENTFORMERCITE & 1.59 & 3.14 & 1.30 & 2.47 \\
        MUSE-VAE\cite{lee2022muse} & 1.38 & 2.90 & 1.09 & 2.10 \\
        
        \midrule
        \EVMODEL*\EVCITE & 1.46 & 3.18 & 1.15 & 2.37 \\
        \EVMODEL-SC & 1.44 & 3.10 & 1.13 & 2.30 \\
        
        \bottomrule
    \end{tabular}
    \caption{
        Comparisons on nuScenes under \emph{best-of-5} and \emph{best-of-10} in meters.
        Lower ADE and FDE indicate better prediction performance.
        Models with ``*'' are reproduced under the same settings.
    }
    \label{tab_nuscenes}
\end{table}

\subsection{Dataset Configurations}

The \textbf{nuScenes}\cite{caesar2020nuscenes,caesar2019nuscenes} is a large-scale real-world dataset of 1000 driving scenes collected in the urban cities of Boston and Singapore.
Each scene has 20 seconds and is annotated at 2 fps.
850 scenes were manually annotated for 23 classes, such as pedestrians and vehicles, and included visibility, activity, and pose attributes.
Note that only vehicles' 2D trajectories $\left\{\mathbf{p}^i_t\right\}_{i, t} = \left\{\left(x^i_t, y^i_t\right)\right\}_{i, t}$ are used in this paper.
Following the settings of \cite{lee2022muse}, we predict future $t_f = 12$ frames' trajectories according to vehicles' past $t_h = 4$ frames' observed trajectories.
The sample interval between two adjacent frames is set to $\Delta t = 0.5s$.
Since the annotations of the official 150 test sets are not available, following previous works like \cite{saadatnejad2022pedestrian}, we use 550 scenes to train, 150 scenes to validate, and the other 150 scenes to test.

\subsection{Baselines}

We choose \TRAJECTRONPP\TRAJECTRONPPCITE, \YNET\YNETCITE, \AGENTFORMER\AGENTFORMERCITE, MUSE\cite{lee2022muse}, and \EVMODEL* as our baselines on nuScenes.

\subsection{Metrics}

Following previous works like \cite{lee2022muse}, we use both \emph{best-of-5} and \emph{best-of-10} validations to evaluate models' performance on nuScenes.
Like the main paper, we denote these metrics as minADE$_{5}$/minFDE$_{5}$ and minADE$_{10}$/minFDE$_{10}$ (short for ADE$_{5}$/FDE$_{5}$ and ADE$_{10}$/FDE$_{10}$).

\subsection{Quantitative Analyses}

\TABLE{tab_nuscenes} reports the quantitative performance of several baseline models and the corresponding \EVMODEL~model.
Although the base model (\EVMODEL) is not specifically designed to predict trajectories in traffic scenes, \MODEL~still shows its capability to model interactions among vehicles.
Compared to the vanilla \EVMODEL, \EVMODEL-SC has a 1.4\% better ADE$_5$ and a 2.5\% better FDE$_5$.
The performance gain brought by the \MODEL~is more remarkable as the number of predicted trajectories rises from 5 to 10, including 1.7\% on the ADE$_{10}$ and 3.0\% on the FDE$_{10}$.

Although \MODEL~could help the base model \EVMODEL~to perform better, there are still noticeable differences in the performance between \EVMODEL-SC and the MUSE-VAE that focus mainly on vehicle trajectory prediction, including 3.7\% and 9.5\% worse ADE$_{10}$ and FDE$_{10}$.
It is worth noting that MUSE-VAE uses additional lane information to help predict better, whereas neither the base model \EVMODEL~nor the corresponding \MODEL~model \EVMODEL-SC do not.
This further inspires us to design meta components for the \MODEL~in traffic prediction scenarios.

\section{Additional Experimental Analyses on the Number of \MODEL~Partitions}
\label{section_appendix_partitions}

\begin{table}[tbp]
    \small
    \centering
    \begin{tabular}{c|c|cc}
        \toprule
        Variations
        & $N_\theta$
        & ADE/FDE
        & Gain (\%) \\

        \midrule
        \VMODEL* & - & 7.04/10.94 & -4.92\%/-2.63\% \\
        \VMODEL-SC-a4 & 1 & 6.96/11.05 & -3.73\%/-3.66\% \\
        \VMODEL-SC-a5 & 4 & 6.79/10.80 & -1.19\%/-1.31\% \\
        \VMODEL-SC & 8 & 6.71/10.66 & (base) \\
        \VMODEL-SC-a6 & 12 & 6.65/10.60 & +0.89\%/+0.56\%\\
        \VMODEL-SC-a7 & 16 & 6.68/10.65 & +0.45\%/+0.09\%\\
        \VMODEL-SC-a8 & 36 & 6.64/10.64 & +1.04\%/+0.19\%\\

        \midrule
        \EVMODEL* & - & 6.73/10.75 & -2.91\%/-3.76\% \\
        \EVMODEL-SC-a4 & 1 & 6.66/10.70 & -1.83\%/-3.28\% \\
        \EVMODEL-SC-a5 & 4 & 6.61/10.55 & -1.07\%/-1.83\% \\
        \EVMODEL-SC & 8 & 6.54/10.36 & (base) \\
        \EVMODEL-SC-a6 & 12 & 6.50/10.34 & +0.61\%/+0.19\% \\
        \EVMODEL-SC-a7 & 16 & 6.46/10.22 & +1.22\%/+1.35\% \\
        \EVMODEL-SC-a8 & 36 & 6.57/10.41 & -0.46\%/-0.48\% \\
        
        \bottomrule
    \end{tabular}
    \caption{
        Ablation studies on verifying the number of \MODEL~partitions $N_\theta$ with different backbone models on SDD.
        Values in the ``Gain'' column are the percentage ADE and FDE gain compared to the base 8-partition model (denoted with ``(base)'').
    }
    \label{tab_ab_partitions}
\end{table}

\begin{figure}[t]
    \centering
    \includegraphics[width=1.0\linewidth]{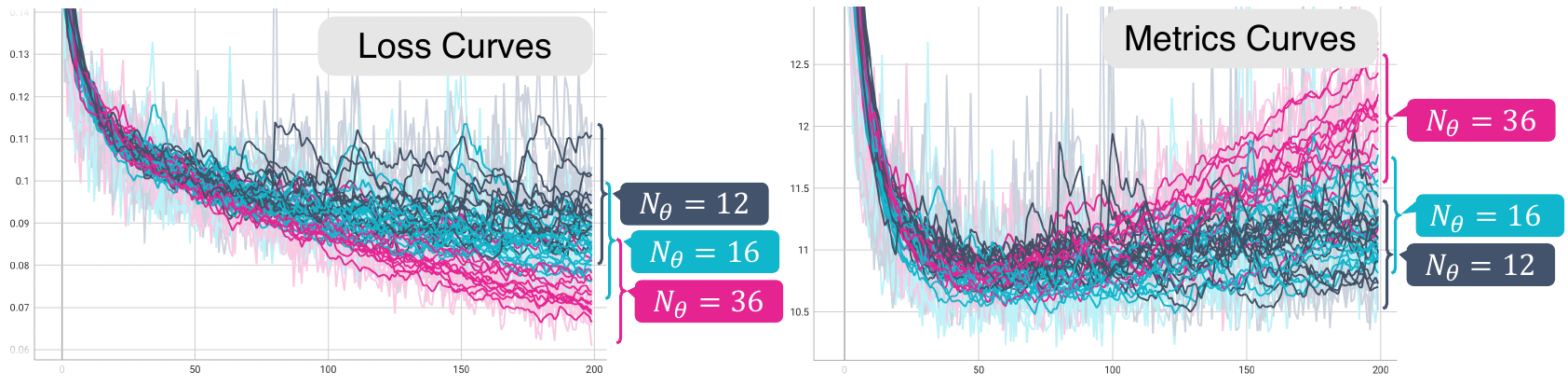}
    \caption{
        Loss curves (left, $\ell_2$ loss) and metrics curves (right, ADE) of \EVMODEL-SC variations a6 to a8 ($N_\theta \in \{12, 16, 36\}$).
    }
    \label{fig_loss_partitions}
\end{figure}

\begin{figure*}[t]
    \centering
    \includegraphics[width=1.0\linewidth]{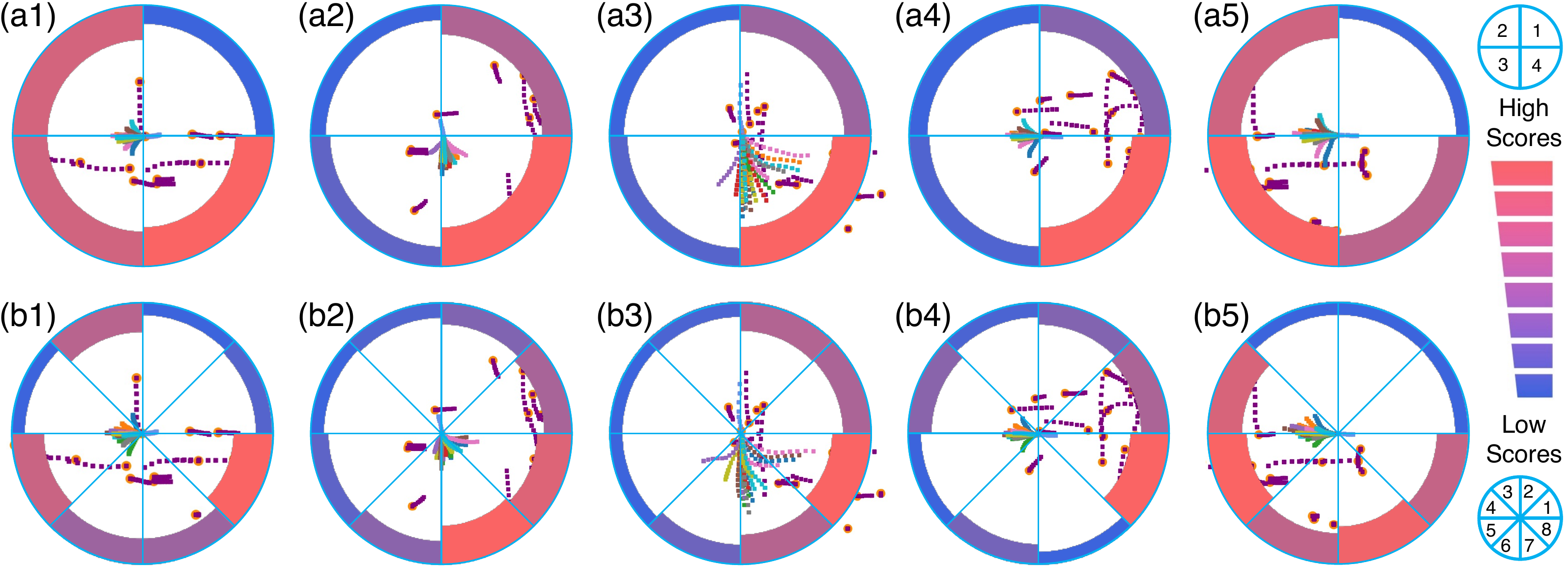}
    \caption{
        Visualized predicted trajectories and the corresponding attention scores of several real-world prediction cases on SDD-little0 provided by the \textbf{4-partition} \EVMODEL-SC (a1) to (a5) and the \textbf{8-partition} \EVMODEL-SC (b1) to (b5).
    }
    \label{fig_attention_partitions}
\end{figure*}

\begin{figure*}[t]
    \centering
    \includegraphics[width=0.916\linewidth]{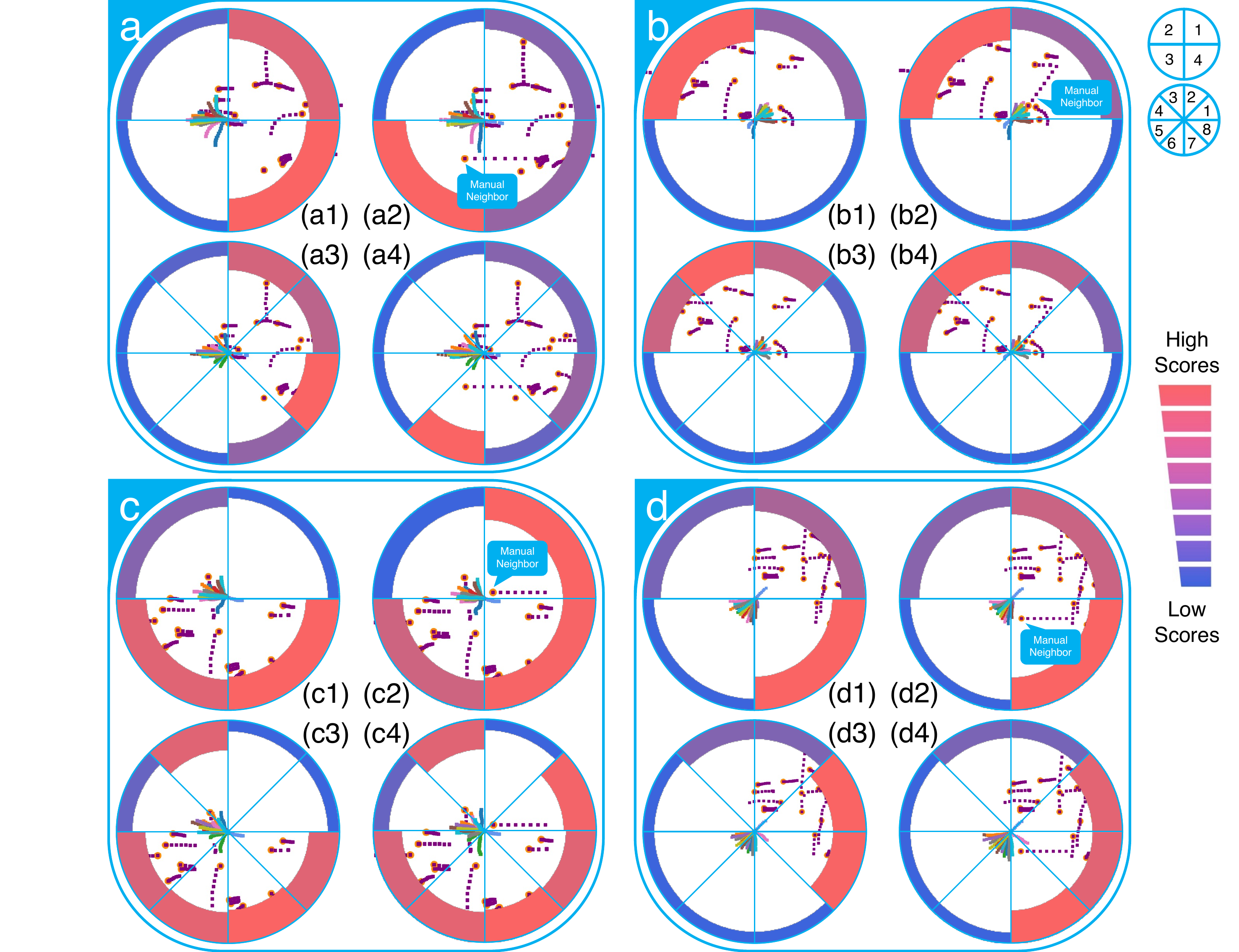}
    \caption{
        Visualized predicted trajectories and the corresponding attention scores of several real-world cases by adding additional manual neighbors.
        For each case $x \in \{\textrm{a, b, c, d}\}$, subfigure ($x$1) is the \textbf{4-partition} ($N_\theta = 4$) model's prediction, and ($x$3) is \textbf{8-partition} ($N_\theta = 8$) prediction.
        subfigures ($x$2) and ($x$4) are obtained by adding manual neighbors to cases ($x$1) and ($x$3), respectively.
    }
    \label{fig_attention_partitions_toy}
\end{figure*}

\subsection{Quantitative Analyses}

We run ablation experiments to validate how the number of \MODEL~partitions $N_\theta$ affects models' quantitative performance.
In \TABLE{tab_ab_partitions}, 8-partition \MODEL~models perform the best, outperforming 4-partition variations for about 1.1\% to 1.8\% ADE and FDE.
Especially, models with $N_\theta = 1$ work even worse, including up to 2.5\% ADE drop compared to 4-partitions'.
Comparing \VMODEL~and \VMODEL-SC-a4, we find that the latter one even has about 0.1 pixels worse FDE.
It aligns with our intuition that the more partitions the higher resolutions for describing social behaviors.
While vice versa, too few partitions may lead to a coarse description of interactions, even mislead the model, thus significantly reducing prediction performance.

Note that due to the settings of predicting trajectories based on 8 historical observed frames on SDD, the maximum number of partitions is set to 8 to prevent unnecessary zero-paddings in trajectories' representations from pulling down the performance of the original backbone trajectory prediction network.
To verify this thought, we expand the \MODEL~to make it available to handle $N_\theta > t_h$ cases by zero-padding trajectory representations (\IE, the $\mathbf{f}^i_{\rm{traj}}$ in Eq. (13)).
Results of variations with postfixes \{a6, a7, a8\} reported in \TABLE{tab_ab_partitions} are obtained under this new setting.
In addition, we have attached the loss curves and metrics curves of these $N_\theta > t_h$ variations in \FIG{fig_loss_partitions}.
It shows that the loss may drop faster as the $N_\theta$ raises, but simultaneously exacerbates the risk of overfitting.
We can further infer that even though a higher $N_\theta$ may provide better results, it also compresses the information in trajectories while reducing training stability.
On balance, $N_\theta = 8$ may be a good compromise (ETH-UCY and SDD).
As a result, we regard that $N_\theta$ should be no more than the $t_h$ in the main paper.

\subsection{Qualitative Analyses}

\FIG{fig_attention_partitions} provides the visualized attention scores in different prediction cases on SDD-little0 with the $N_\theta = 4$ (subfigures (a1) to (a5)) and the $N_\theta = 8$ ((b1) to (b5)) \EVMODEL-SC models.
These two models are trained and validated under the same condition except for the $N_\theta$.

Comparing \FIG{fig_attention_partitions} (a3) and (b3), the 8-partition model provides trajectories with different social behaviors for $\theta \in \left[1.5\pi, 2\pi\right)$, \IE, partitions 7 and 8.
In detail, predictions in partition-8 mostly try to avoid the right-coming neighbor, while predictions in partition-7 mostly walk as normal cases.
For the 4-partition model's predictions in \FIG{fig_attention_partitions} (a3), predictions within the whole partition-4 all present the avoidance tendance, even though some predicted trajectories are far away from the existing neighbors.
Similar cases also appear in cases (a2, partition-4) v.s. (b2, partitions 7 and 8) and cases (a5, partition-3) v.s. (b5, partitions 5 and 6).
All these comparisons point out that a smaller number of \MODEL~partitions may lead to a coarser recognition and modeling of social behaviors, thus further causing misleading shifts in the predicted trajectories.

We also add manual neighbors to real-world prediction cases on SDD-little0 to validate both $N_\theta = 4$ and $N_\theta = 8$ \EVMODEL-SC models' responses.
As shown in \FIG{fig_attention_partitions_toy}, $N_\theta = 8$ model presents better spatial resolutions for handling social interactions.
For example, compared to the $N_\theta = 4$ case (c2, partition-1), the corresponding $N_\theta = 8$ partition (c4, partition-2) has been less affected due to the manual neighbor.
As a result, predictions in 8-partitions cases \{(c4, partition-3), (c4, partition-4)\} show different interactive trends.
These results indicate that 8-partition \MODEL~models have better angular resolution to model potential social interactions as well as quantify their roles in modifying forecast results.

\begin{table}[tp]
    \small
    \centering
    \begin{tabular}{c|ccc}
        \toprule
        Models & \makecell[c]{ADE/FDE $\downarrow$ \\ (ETH-UCY)} & Time $\downarrow$ & Paras. $\downarrow$ \\

        \midrule
        \SOCIALLSTM\SOCIALLSTMCITE & 0.72/1.54 & 1180 ms & 264K \\
        \SRLSTM\SRLSTMCITE & 0.45/0.94 & 1179 ms & 64.9K \\
        \PECNET\PECNETCITE & 0.29/0.48 & 607 ms & 2.10M \\
        \PEEKING\PEEKINGCITE & 0.46/1.00 & 114 ms & 360.3K \\
        \SOCIALGAN\SOCIALGANCITE & 0.58/1.18 & 97 ms & 46.3K \\
        \DAGNET\DAGNETCITE & \NA & 46 ms & 2.35M \\
        \STGCNN\STGCNNCITE & 0.44/0.75 & 2.0 ms & 7.6K \\
        \STCNET\STCNETCITE & 0.38/0.68 & 1.3 ms & \MARK{0.7K} \\
        \VMODEL*\VCITE & 0.18/0.28 & 19 ms & 1.91M \\
        \EVMODEL*\EVCITE & 0.17/0.28 & 21 ms & 1.92M \\

        \midrule
        \VMODEL-SC & 0.17/0.27 & 23 ms & 1.92M \\
        \EVMODEL-SC & 0.17/0.27 & 24 ms & 1.98M \\

        \bottomrule
    \end{tabular}
    \caption{
        Comparisons of inference time and model parameters.
        Results are obtained from \cite{li2021spatial} on one NVIDIA GeForce GTX 1080Ti card.
        Models with ``*'' are reproduced with PyTorch.
    }
    \label{tab_parameter}
\end{table}

\begin{table}[htb]
    \small
    \centering
    \begin{tabular}{c|ccccc|c}
        \toprule
        \multirow{2}{*}{Model} &
        \multicolumn{5}{c|}{Inference time @batchsize} &
        \multirow{2}{*}{Parameters} \\

        & 1 & 50 & 100 & 500 & 1000 & \\

        \midrule
        \VMODEL & 28 & 30 & 31 & 38 & 81 & 1,911,264 \\
        \VMODEL-SC & 34 & 35 & 36 & 55 & 88 & 1,923,936 \\

        \midrule
        \EVMODEL & 28 & 33 & 37 & 67 & 112 & 1,976,864 \\
        \EVMODEL-SC & 34 & 39 & 43 & 73 & 119 & 1,989,536 \\

        \bottomrule
    \end{tabular}
    \caption{
        Inference times (in milliseconds) at different batch size settings (from 1 to 1000) and the number of trainable parameters of \VMODEL, \EVMODEL, and their corresponding \MODEL~models.
        Results are obtained by running models (PyTorch) on one Apple Mac mini (M1, 2020) with 8GB memory.
    }
    \label{tab_speed}
\end{table}

\section{Parameters and Inference Times}

\textbf{Comparisons with Other Baselines.}
We compare the inference speed and the number of parameters of different models in \TABLE{tab_parameter}.
All results are measured on one NVIDIA GeForce GTX 1080Ti GPU (short for ``1080Ti'').
Since the official codes of \VMODEL~and \EVMODEL~are implemented with TensorFlow and run slowly in our Python environment on the server, we reproduce their codes with PyTorch and report their running time (batch size is set to 1, marked with ``*'') in \TABLE{tab_parameter}.
From these results we can see that the \MODEL~itself would not lead to a large number of computations and extra trainable variables.
Compared to the original models, the inference times of their corresponding \MODEL~models are still considerable.

~\\\textbf{Further Discussions on the Inference Speed.}
Considering that the platform on which trajectory prediction models are running may not be equipped with high-performance computing devices, all results reported in \TABLE{tab_speed} are obtained on one Apple Mac Mini with an Apple M1 chip (8GB memory), which performs similarly to current iPhones and iPads.
Additionally, several researchers like \cite{li2021spatial} have defined the \emph{low-latency trajectory prediction}, which indicates that the trajectory prediction method should predict trajectories within the sampling interval to achieve the real-time prediction goal.
For example, when predicting trajectories on ETH-UCY with a sample rate of 2.5 fps, the implementing time of the model should be less than 400 ms.
Results in \TABLE{tab_speed} show that the proposed methods could meet the low-latency standard even when running on the Apple M1 chip, indicating their potential to be applied to complex application scenarios.

\section{Additional Visualized Toy Examples}

\begin{figure*}[t]
    \centering
    \includegraphics[width=1.0\linewidth]{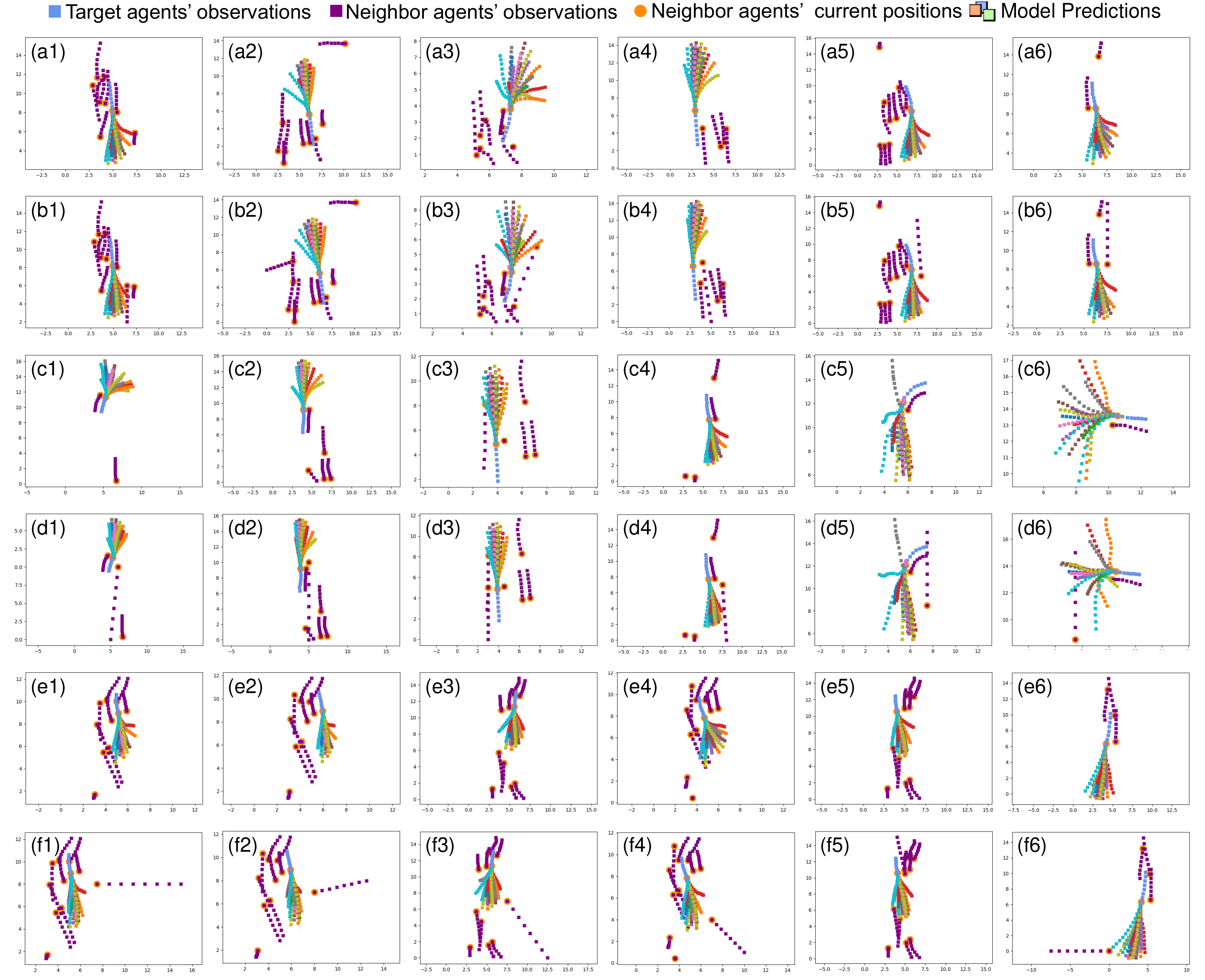}
    \caption{
        Toy examples (linear interpolation) on validating the effectiveness of the overall modification of social interactions.
        We add manual neighbors to the original ETH-UCY prediction scenes and visualize how they change the predicted trajectories.
        Prediction case in subfigure ($xn$), where $x \in \{\textrm{a, c, e}\}, n\in\{1, 2, 3, 4, 5, 6\}$, represents the original prediction scene in UCY-zara1, and the corresponding ($yn, y \in \{\textrm{b, d, f}\}$) case represents prediction considering the manual neighbor.
    }
    \label{fig_factor_vis}
\end{figure*}

\begin{figure*}[t]
    \centering
    \includegraphics[width=1.0\linewidth]{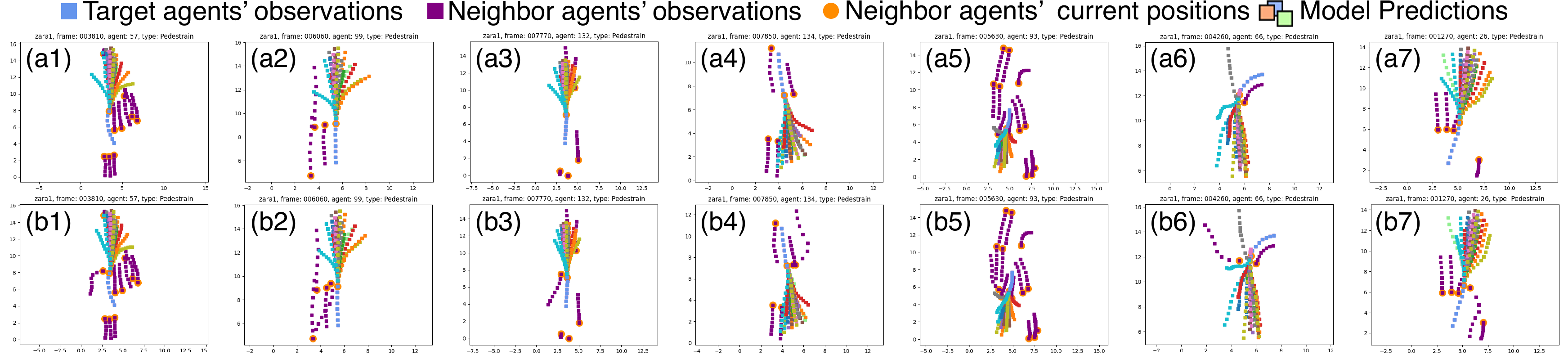}
    \caption{
        Toy example (linear-velocity interpolation) on validating social interactions.
        Compared to the linearly-interpolated trajectories, we add several non-linear patterns to the trajectories of manual neighbors to further reflect their fine-level motions. 
        The prediction case in subfigure (a$n$), where $n\in\{1, 2, 3, 4, 5, 6, 7\}$, represents the original prediction scene in UCY-zara1, and the corresponding (b$n$) case represents prediction considering the curved-moving manual neighbor.
    }
    \label{fig_curve_neighbor}
\end{figure*}

To demonstrate the effectiveness of the proposed \MODEL~in handling different social interaction cases, following the settings in Section 4.3 \textbf{Toy Examples I (Social Interactions)}, we provide more visualized toy examples in the real-world UCY-zara1 prediction scenes in this section.
In these toy examples, we add one manual neighbor to each prediction case, thus visualizing how \MODEL~modifies the original predicted trajectories under different interaction contexts. 

In the main paper, we use a simple linear interpolation method to simulate manual neighbors' trajectories.
For agent $i$, given two points $\mathbf{p}^i_0$ and $\mathbf{p}^i_{t_h}~(1 \leq t \leq t_h)$, the linearly-interpolated coordinate $\mathbf{p}^i_t$ is computed via
\begin{equation}
    \mathbf{p}^i_t = \mathbf{p}^i_0 + \frac{
        \mathbf{p}^i_{t_h} - \mathbf{p}^i_0}{
            t_h
        } t.
\end{equation}

\FIG{fig_factor_vis} includes more visualized predictions under different linearly interpolated manual neighbor settings.
We also designed a non-linear interpolation method to further validate \MODEL's capability, which linearly interpolates the velocity from each adjacent two of the three given points to generate manual neighbors with curved trajectories via
\begin{align}
    \mathbf{v}^i_t &= \mathbf{p}^i_t - \mathbf{p}^i_{t-1},\\
    \mathbf{v}^i_t &= \mathbf{v}^i_0 + t \Delta \mathbf{v},\\
    \sum_{t = 1}^{t_h} \mathbf{v}^i_t &= \mathbf{p}^i_{t_h} - \mathbf{p}^i_0.
\end{align}
Thus, $\Delta \mathbf{v}$ can be represented as
\begin{align}
    \Delta \mathbf{v} &= \frac{2 (\mathbf{p}^i_{t_h} - \mathbf{p}^i_0 - \mathbf{v}^i_0 t_h)}{t_h (t_h + 1)},
\end{align}
and we can finally determine the coordinate $\mathbf{p}^i_t$ at any moment $t$.
Formally,
\begin{equation}
    \mathbf{p}^i_t = \mathbf{p}^i_0 + \sum_{n=1}^{t} n \Delta \mathbf{v}.
\end{equation}
These trajectories and the corresponding \MODEL~predictions are shown in \FIG{fig_curve_neighbor}. 
In both figures, we observe that after adding manual neighbors with a certain velocity around the target agent, its new predicted trajectories tend to keep a certain \emph{social distance} to the manual neighbor in most cases.
For example, in \FIG{fig_factor_vis} (b2, b3, b4) and \FIG{fig_curve_neighbor} (b1, b3, b4), the target agents are predicted to move away from the manual neighbors dramatically.
In some cases, like \FIG{fig_factor_vis} (d6, f1) and \FIG{fig_curve_neighbor} (b5, b7), the originally predicted trajectories of the target agent before adding the manual neighbor have already demonstrated a strong trend of movement toward certain destinations. 
Among these cases, if we add a manual neighbor that also moves toward such a destination with a relatively fast velocity, the newly predicted trajectories of the target agent may change heavily to avoid possible collisions or keep certain social distances with the manual neighbor.

Unlike these situations, \FIG{fig_factor_vis} (f6) and \FIG{fig_curve_neighbor} (b2), represent a different way to handle interactions in which the predicted trajectories have shifted to the left to avoid the fast-moving manual neighbor coming from the left side, rather than shifted to the right side.
These phenomena demonstrate that the \MODEL~models could dynamically handle different interactive contexts in different prediction scenes, thus providing trajectories in line with social rules.
In short, the three meta components (velocity, distance and direction) used in \MODEL~have the potential to reflect different interactive contexts and further promote the prediction networks to learn to generate divergent trajectories.

However, we also observe that there exist some cases in which predictions do not comply with interactive contexts.
In \FIG{fig_factor_vis} (d3), \MODEL~model still remains the way it forecasts trajectories for the target agent even after adding a near enough manual neighbor with a relatively fast velocity.
In \FIG{fig_factor_vis} (d2), after adding the fast-moving manual neighbor on the right side, the left part of the predicted trajectories are pruned off.
Although the quantitative prediction performance has not been influenced, it actually constrains the diversity of the predicted trajectories.
Therefore, the three meta components (velocity, distance and direction) used in \MODEL~are still worthy of further studies to simulate and forecast in more complex interactive cases.

\section{Further Discussions on Limitations}
\label{section_appendix_limitations}

\begin{figure*}[t]
    \centering
    \includegraphics[width=1.0\linewidth]{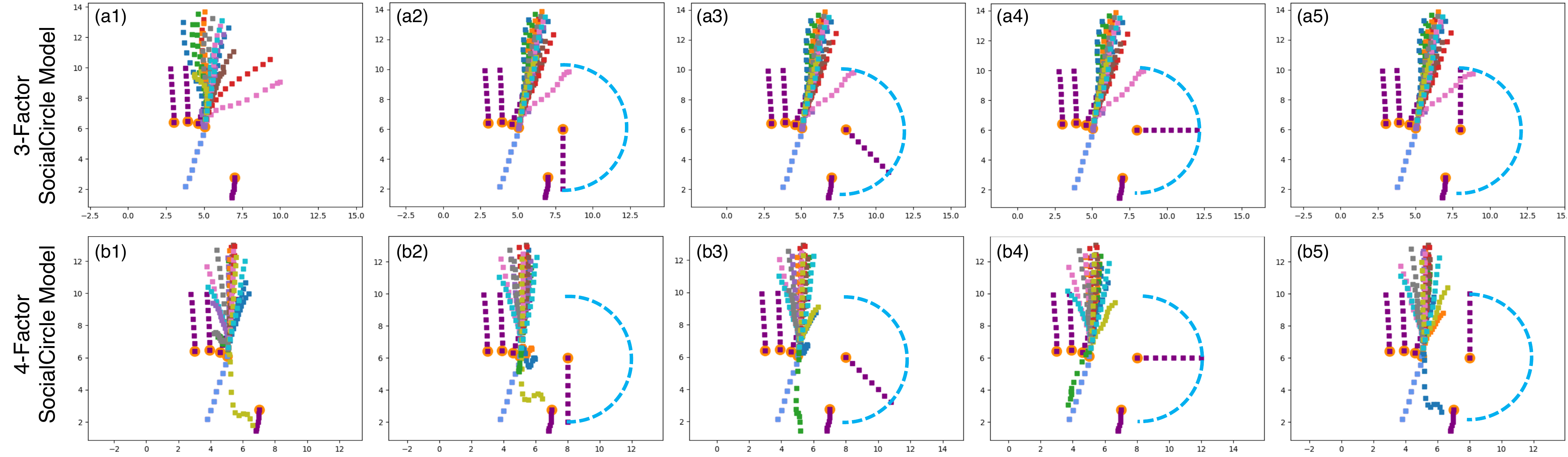}
    \caption{
        Visualized \EVMODEL-SC predictions with manual neighbors with different movement directions.
        In this toy experiment, we set $d_m = 2.97$ and $v_m = 4.00$.
        (a1) to (a5) are predictions provided by the \textbf{3-factor} \MODEL~model, and (b1) to (b5) are predictions by \textbf{4-factor} model.
        Cases (a1) and (a5) are their original predictions without any given manual neighbors.
    }
    \label{fig_limitations}
\end{figure*}

\begin{table}[tbp]
    \small
    \centering
    \begin{tabular}{c|c|cc}
        \toprule
        Variations & V D R mR & ADE/FDE & Drop (\%) \\

        \midrule
        \EVMODEL* & \xm \xm \xm~~\xm & 6.73/10.75 & -2.91\%/-3.76\% \\
        \EVMODEL-SC & \cm \cm \cm~~\xm & 6.54/10.36 & (base) \\
        \EVMODEL-SC-4f & \cm \cm \cm~~\cm & 6.84/10.94 & -4.59\%/-5.60\% \\
        
        \bottomrule
    \end{tabular}
    \caption{
        Ablation studies on validating the movement direction (``mR'') factor on SDD.
        ``V'', ``D'', and ``R'' represent current velocity, distance, and direction factors.
        Values in ``Drop'' are the percentage matrices drop compared to the base model.
    }
    \label{tab_ab_4factors}
\end{table}

\begin{figure*}[t]
    \centering
    \includegraphics[width=1.0\linewidth]{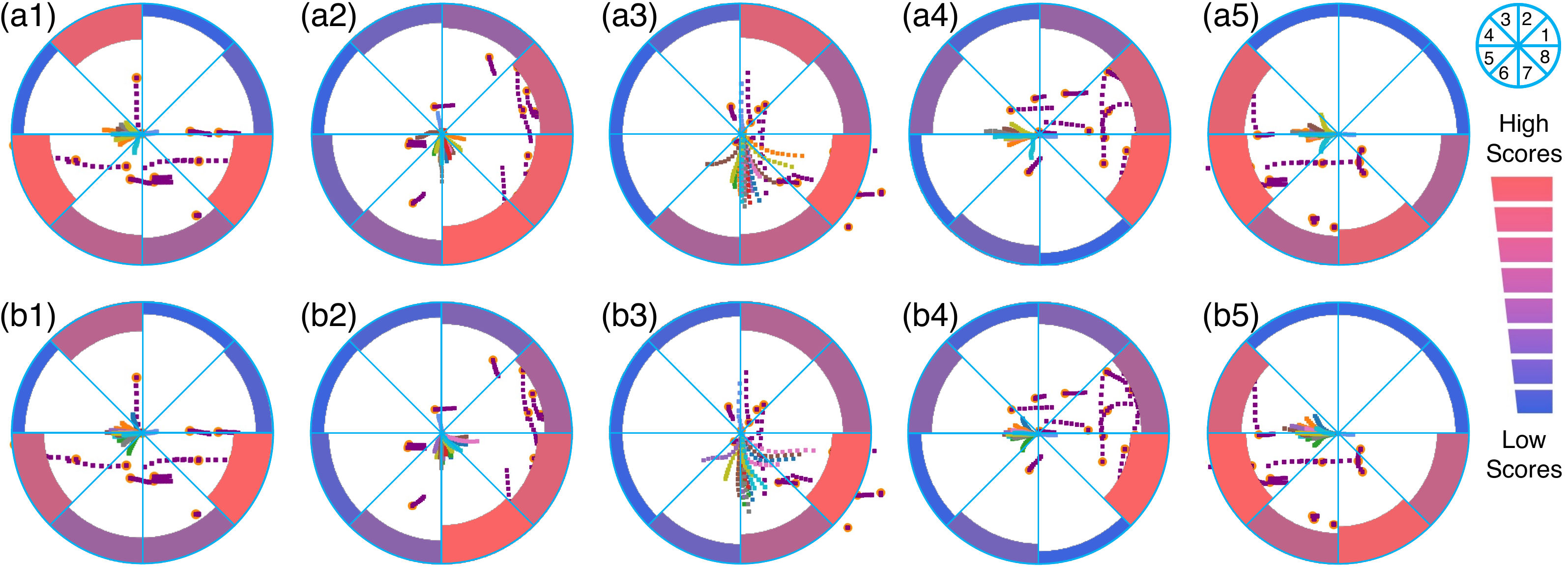}
    \caption{
        Visualized predicted trajectories and their corresponding attention scores in several real-world prediction cases (SDD-little0) provided by the \textbf{4-factor} \EVMODEL-SC (a1) to (a5) and the \textbf{3-factor} \EVMODEL-SC (b1) to (b5).
    }
    \label{fig_attention_factors}
\end{figure*}

\begin{figure*}[t]
    \centering
    \includegraphics[width=0.916\linewidth]{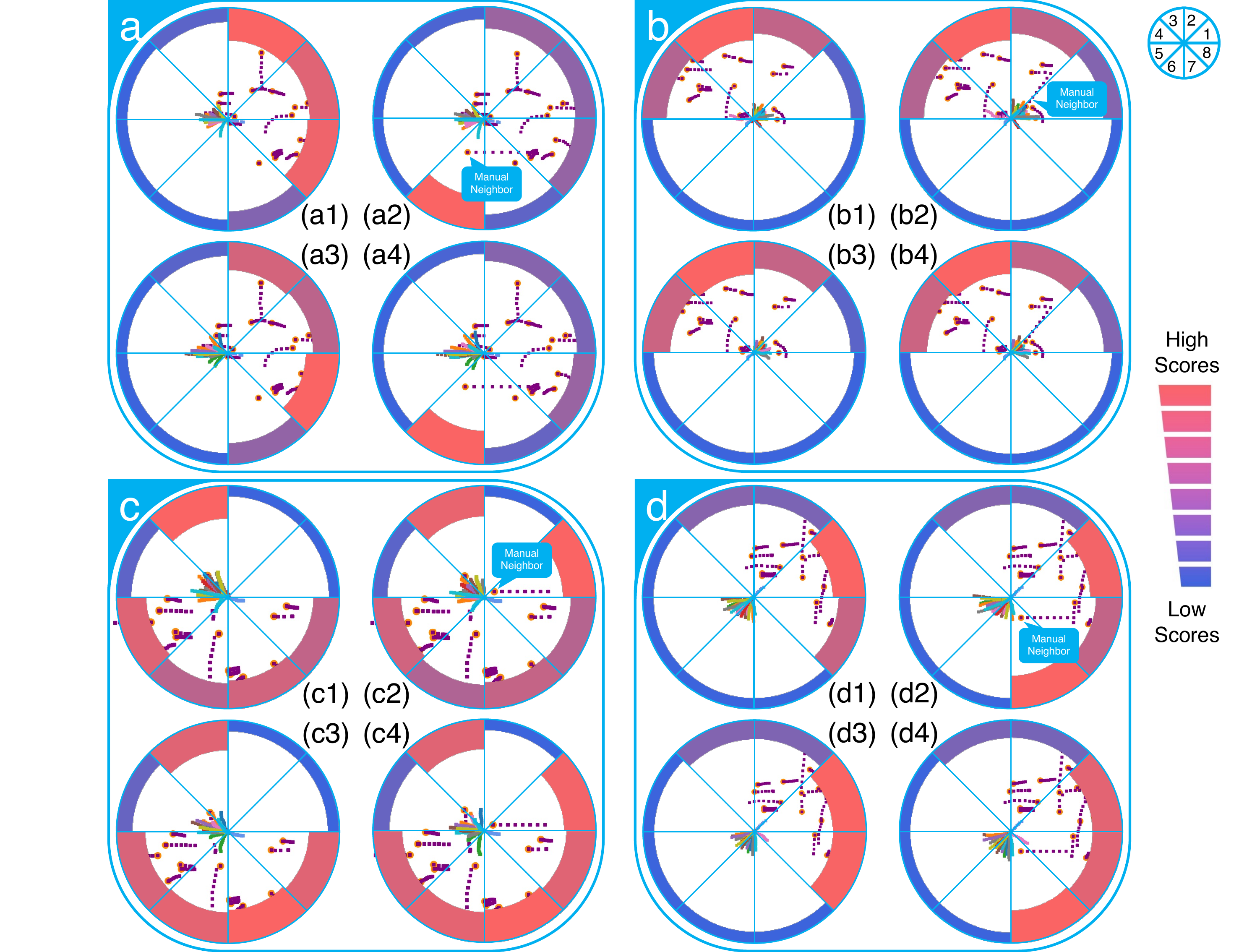}
    \caption{
        Visualized predicted trajectories and the corresponding attention scores of several real-world cases by adding additional manual neighbors.
        For each case $x \in \{\textrm{a, b, c, d}\}$, subfigure ($x$1) is the \textbf{4-factor} model's prediction, and ($x$3) is the \textbf{3-factor} model's prediction.
        subfigures ($x$2) and ($x$4) are obtained by adding manual neighbors to cases ($x$1) and ($x$3), respectively.
    }
    \label{fig_attention_factors_toy}
\end{figure*}

As mentioned in the ``Limitations'' section, neighbor agents' movement directions have not been considered in the proposed \MODEL.
This section further discusses whether the movement direction factor should be considered as one of the \MODEL~meta components.

\subsection{Limitation Analysis}

As shown in \FIG{fig_limitations}, we conducted another toy experiment to show models' responses to the manual agent with different movement directions.
In all 3-factor cases (a2) to (a5), the \MODEL~model forecasts almost the same trajectories (except for the noise factor for random generation).
It is worth noting that the predictions in case (a3) are relatively ``dangerous'', for there might be potential collisions or too-close social distances with the manual neighbor.

From the point of view of network training, we can simply understand that the whole prediction network forecasts an ``average'' trajectory to satisfy all these training samples with the same \MODEL~but move in different directions.
As a result, it may predict trajectories with avoidances for the neighbors that may not collide with the target agent (like \FIG{fig_limitations} (a5)), or may still collide with others (like \FIG{fig_limitations} (a3)).

It should be noted that these extreme cases in the toy experiments are rarely seen in real-world prediction scenarios.
In most ETH-UCY and SDD scenes, \MODEL~models still work as expected.
Nevertheless, these few uncovered social interaction cases still indicate their limitations, although they have achieved better quantitative performance.

\subsection{The Movement Direction Factor.}

Following the ``lite-rules'' assumption, we attempt to add the movement direction factor to provide detailed interactive information.
It is defined as the average of each neighbor's moving direction located in some partition.
Formally,
\begin{equation}
    \mathbf{f}^i_{\rm{mdir}} \left(\theta_n\right) =
        \frac{1}{\left| \mathbf{N}^i(\theta_n) \right|}
        \sum_{j \in \mathbf{N}^i(\theta_n)}
        {\rm{atan2}} \left( f_{\rm{2D}}\left(
            \mathbf{p}^j_{t_h} -
            \mathbf{p}^j_{1}
        \right)\right).
\end{equation}
The corresponding 4-factor \MODEL~meta vector is
\begin{equation}
    \mathbf{f}^i_{\rm{meta}} \left(\theta_n\right) = \left(
        \mathbf{f}^i_{\rm{vel}} \left(\theta_n\right),
        \mathbf{f}^i_{\rm{dis}} \left(\theta_n\right),
        \mathbf{f}^i_{\rm{dir}} \left(\theta_n\right),
        \mathbf{f}^i_{\rm{mdir}} \left(\theta_n\right)
    \right)^\top.
\end{equation}

\subsection{Ablation Studies and Visualized Analyses of the Movement Direction Factor}

\noindent\textbf{Quantitative Analyses.}
We run experiments to quantitatively validate the usefulness of this movement direction factor on SDD, and their results are reported in \TABLE{tab_ab_4factors}.
By adding this additional factor, the \EVMODEL-SC-4f's performance drops significantly.
Compared to the 3-factor \EVMODEL-SC, it has 4.59\% worse ADE and 5.60\% worse FDE.
Especially, its performance is even worse than the non-\MODEL-model \EVMODEL, which means that just adding such a simple new factor prevents other factors from expressing their contributions.

We infer that the movement direction factor brings more complex constraints to each prediction case, thus making the training process more difficult while reducing the model's generalization capability.
In detail, the current three factors (velocity, distance, direction) are relatively ``weak'' rules to describe social interactions.
Thus, the obtained \MODEL s could be similar even in different prediction cases.
On the contrary, the movement direction factor varies from $0$ to $2\pi$ for each neighbor in each partition, which brings extra ``complexity'' for each interactive case, thus further increasing the difficulty of model training in the case of the same network structure and training data.

\textbf{Validation of Moving Directions.}
In \FIG{fig_limitations} (b1) to (b5), we visualize the predicted trajectories provided by the 4-factor \EVMODEL-SC corresponding to cases (a1) to (a5).
We can easily see that predictions in cases (b2) to (b5) are different due to the various moving directions of the given manual neighbor.
However, trajectories forecasted by the 4-factor model are far worse than those predicted by the 3-factor model.
In detail, several randomly generated trajectories are distributed ``messily'' around the target agent, which could be caused by the ``misleading'' of 4-factor \MODEL~on predicted trajectories at different spatial positions.
In other words, the newly added movement direction factor may prevent the backbone prediction model from exhibiting its original prediction performance.

\textbf{Moving Directions and Attention Scores.}
We visualize predictions of both 3-factor and 4-factor \MODEL~models on more real-world scenes in \FIG{fig_attention_factors} and toy prediction cases with manual neighbors in \FIG{fig_attention_factors_toy}.
Comparing \FIG{fig_attention_factors} (a1) and (b1), it shows that more \MODEL~partitions have been paid attention to (red colored partitions) in the 4-factor model in (a1) than (b1).
Cases \{(a2), (b2)\} and \{(a3), (b3)\} also show similar trends.
It means that more partitions or neighbors (\IE, more ``rules'') are considered simultaneously to make final predictions for the 4-factor \MODEL~model.
In addition, predictions provided by the 4-factor \MODEL~could hardly handle interactive behaviors in complex social interaction cases.
For example, predictions in partitions 7 and 8 in \FIG{fig_attention_factors} (b3) show strong avoidance trends to the coming neighbor.
In contrast, predictions in the same partitions in (a3) have almost no responses.
More visualized toy results with manual neighbors on real-world scenes are available in \FIG{fig_attention_factors}.



\subsection{Summary of the Movement Direction Factor}

The 3-factor \MODEL~(velocity, distance, direction) could not reflect neighbor agents' moving directions when modeling social interactions and forecasting trajectories.
It takes an ``average'' way to handle neighbors with different movement directions, which means that its forecasted trajectories may not fit the interaction context well in some ``extreme'' interaction cases (like \FIG{fig_limitations} (a3)).

We try to address this limitation by adding the new movement direction factor to the \MODEL~meta components.
However, the newly added factor may lead to a performance drop.
As we can see from the visualized predictions and attention scores, it is most likely due to adding too many constraints to the interaction cases, which reduces the model's ability to generalize across different complex prediction scenarios.
Although the new factor could help to represent better interactive behaviors in some specific cases, degrading the original performance of the prediction model is something we do not expect.
Therefore, the movement direction factor is deprecated in the \MODEL.
The currently proposed \MODEL~is a compromise that devotes itself to describing interactive behaviors through as few rules as possible while maximizing its usability in different trajectory prediction scenes.
We will further investigate this limitation in our subsequent work.


\end{document}